\documentclass[10pt,twocolumn,letterpaper]{article}

\usepackage[pagenumbers]{cvpr} 

\definecolor{cvprblue}{rgb}{0.21,0.49,0.74}
\usepackage[pagebackref,breaklinks,colorlinks,allcolors=cvprblue]{hyperref}
\usepackage{comment}
\usepackage{booktabs} 
\usepackage{multirow}

\usepackage{booktabs,tabularx,array,multirow}
\usepackage[table]{xcolor}
\usepackage{amssymb}

\definecolor{RoboBlue}{RGB}{235,244,250}
\definecolor{RoboGray}{RGB}{248,249,251}

\definecolor{RoboMuted}{RGB}{103,113,125}

\newcommand{\srci}[2]{%
    \shortstack[c]{%
        #1\\[-1pt]
        {\scriptsize\textcolor{RoboMuted}{[#2]}}%
    }%
}

\usepackage{tabularx}
\usepackage{array}
\usepackage{amssymb}
\usepackage{fontawesome5}

\def\confName{CVPR}
\def\confYear{2027}
\newcommand{\methodname}{Robo-EMG}

\usepackage{booktabs,tabularx,array}
\usepackage{xcolor,colortbl}

\definecolor{RoboBlue}{RGB}{235,244,250}
\definecolor{RoboMuted}{RGB}{103,113,125}

\usepackage{graphicx}

\newcommand{\tasklabel}[2]{%
    \raisebox{-0.33\height}{%
        \includegraphics[
            height=4.6mm
        ]{sec/figures/task_icons/#1.pdf}%
    }%
    \hspace{0.45em}#2%
}

\title{Robo-EMG: A Teleoperator EMG Dataset for\\
Learning Robot Manipulation Policies}

\author{
Anupam Pani \quad Wenbo Li \quad Jiahao Zhang \quad Yuxian Li\\
Huang Jiangzixuan \quad Agastya Krish Baliga$^{*}$ \quad Yanchao Yang\\[4pt]
The University of Hong Kong (HKU)\\
{\small $^{*}$Work done during an internship at HKU.}
}

\begin{document}
\maketitle
\begin{abstract}
Vision-language-action (VLA) policies learn
from robot observations and demonstrated
actions, but typically overlook the muscle
activity of the human operator during
teleoperation. We investigate whether
electromyography (EMG) can provide
complementary supervision for VLA learning
without requiring physiological measurements
at inference. We introduce \textbf{Robo-EMG},
a dataset of 5,500 human-teleoperated
demonstrations across 35 real-world and
simulated manipulation tasks, pairing
eight-channel EMG with robot observations
and actions. To identify informative
relationships between muscle activity and
robot behavior, we analyze EMG--action
dependence across temporal windows and
offsets using a consensus of mutual-information
estimators. We then use the aligned EMG
windows to construct learned representations
through a masked spectrogram autoencoder.
An auxiliary prediction head supervises
the VLM's multimodal features by predicting
these EMG representations alongside the
original action-learning objective. Across
nine real-world manipulation tasks, EMG
supervision improves mean success from
40.9\% to 58.9\%. We further observe
improvements under camera occlusion, across
VLA backbones, and in simulated evaluations
under clean and perturbed visual conditions.
These results demonstrate the potential of
human physiological activity as training-only
supervision for robotic manipulation.
\end{abstract}

\section{Introduction}
\label{sec:introduction}

\begin{figure*}[t]
    \centering
    \includegraphics[
        width=\textwidth
    ]{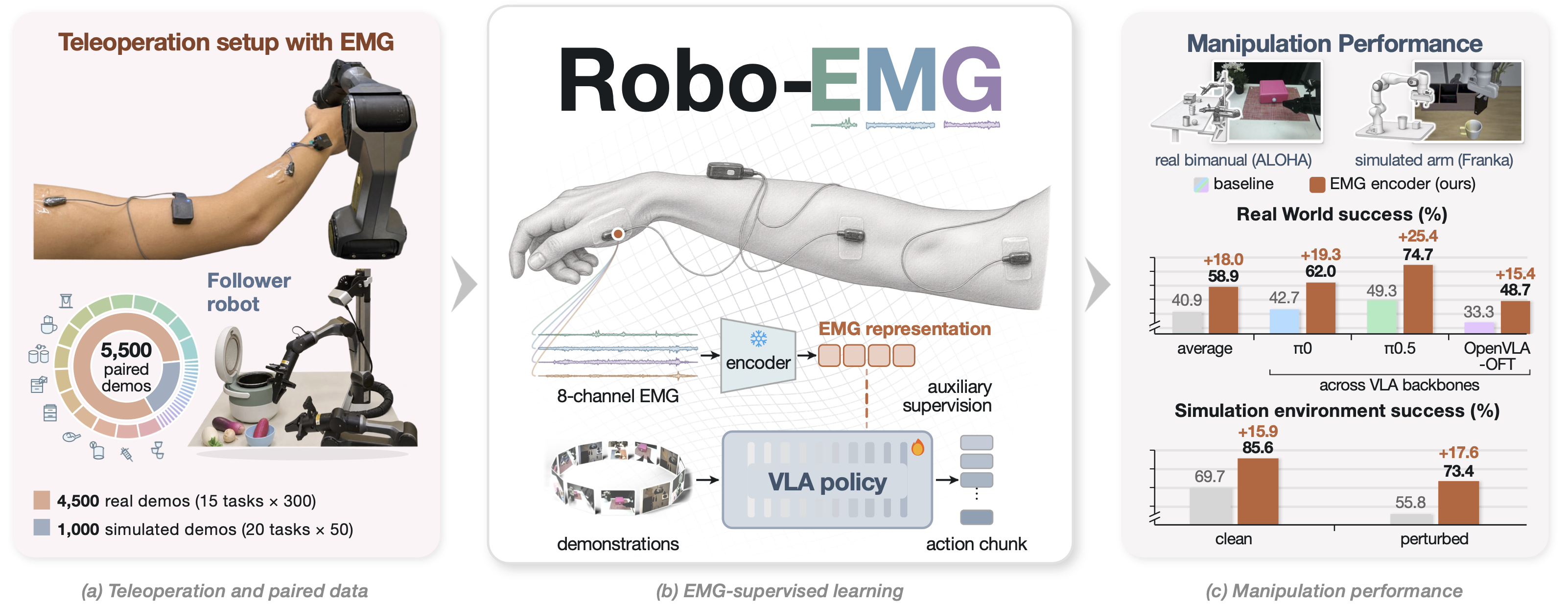}
    \caption{
        \textbf{Overview of Robo-EMG.}
    (a) We collect 5,500 human-teleoperated
    demonstrations paired with eight-channel EMG:
    4,500 real-world and 1,000 simulated trajectories.
    (b) Temporally aligned EMG embeddings provide
    auxiliary supervision during VLA training,
    without requiring EMG at inference.
    (c) EMG supervision improves real-world
    manipulation success across VLA backbones
    and held-out simulation performance under
    clean and perturbed visual conditions.
    }
    \vspace{-4mm}
    \label{fig:teaser}
\end{figure*}

Vision-language-action models have made remarkable progress in robotic manipulation,
learning to translate visual observations, language instructions, and robot states into actions.
Yet, demonstrations collected through human teleoperation tell only part of the story.
They capture what the robot sees and does, but typically overlook the muscle activity of
the human guiding its movements. Behind every grasp, reach, and object manipulation lies
coordinated human muscle activity that is absent from conventional robot demonstrations.
This raises a natural question: are these physiological signals useful, and can we use this missing information
to teach robots better manipulation policies?

Surface electromyography (EMG) offers a non-invasive way to capture electrical activity
associated with muscle excitation. Such activity can precede measurable mechanical
responses ~\cite{Norman1979,Koirala2015}, while the amount of observed EMG history
influences movement recognition and prediction ~\cite{Smith2011,Koch2017}.
These properties make EMG a promising complementary signal to robot actions.
However, muscle activity and robot behavior do not necessarily align at the same
instant, particularly during teleoperation. The challenge is therefore not simply
to record synchronized actions and EMG, but to identify the informative portions of the signal and use
them meaningfully for robot learning.

Motivated by this, we introduce \textbf{Robo-EMG}, a dataset and learning framework
that incorporates human EMG signals into VLA training. We collect 5,500 human-teleoperated
demonstrations spanning 35 manipulation tasks, including 4,500 real-world ALOHA
trajectories and 1,000 simulated environment trajectories. In both environments,
eight-channel EMG is recorded alongside robot observations and actions,
providing a synchronized view of human muscle activity and the resulting robot
behavior. Fig.~\ref{fig:teaser} provides an overview of the dataset, proposed
framework, and manipulation results.

Building on these paired recordings, we investigate how EMG can serve as an
additional source of supervision rather than another input modality.
We first identify informative EMG-action temporal relationships using a
consensus of mutual-information estimators. The aligned EMG windows are then
encoded into learned representations using a masked spectrogram autoencoder.
During VLA training, an auxiliary prediction head encourages the model's
multimodal features to capture information associated with the demonstrated
muscle activity, alongside the original action-learning objective.
Importantly, this supervision is needed only during training: the resulting
policy generates actions using its standard inputs, without requiring EMG
measurements at deployment.
We evaluate Robo-EMG across real-world and simulated manipulation settings.
On nine real-world tasks, EMG supervision improves mean success from 40.9\%
to 58.9\%. The benefits also extend to
camera-view occlusion, different VLA backbones, and simulated manipulation
under clean and visually perturbed conditions. These findings suggest that
human muscle signals can provide
a useful learning signal for robotic manipulation.

Our contributions are threefold:
\textbf{(1)} We introduce \textbf{Robo-EMG}, a dataset of 5,500 human-teleoperated
demonstrations pairing eight-channel EMG with robot observations and actions
across real-world and simulated manipulation tasks.
\textbf{(2)} We develop an MI-guided temporal analysis and learned EMG representations
to construct informative  supervision targets.
\textbf{(3)} We propose a \textbf{training-only EMG supervision framework} and
demonstrate improved manipulation performance across tasks, visual conditions,
and VLA architectures, without requiring EMG at inference.

\section{Related Work}
\label{sec:related_work}

\textbf{Vision-language-action models.}
Vision-language-action models learn robot
control from visual observations, language
instructions, and demonstrated actions.
RT-1~\cite{brohan2023rt1roboticstransformerrealworld} demonstrated scalable
transformer-based robot control, while
RT-2~\cite{brohan2023rt2visionlanguageactionmodelstransfer} incorporated web-scale
vision-language knowledge into action prediction.
Octo~\cite{octomodelteam2024octoopensourcegeneralistrobot} and OpenVLA~\cite{kim2024openvlaopensourcevisionlanguageactionmodel}
advanced generalist robot policies through
pretraining on diverse demonstrations.
More recently, $\pi_0$~\cite{black2026pi0visionlanguageactionflowmodel} and
$\pi_{0.5}$~\cite{intelligence2025pi05visionlanguageactionmodelopenworld} explored flow-based
action generation and heterogeneous co-training,
while OpenVLA-OFT~\cite{kim2025finetuningvisionlanguageactionmodelsoptimizing} improved
VLA fine-tuning and action decoding.
Robo-EMG introduces a complementary source
of supervision using human muscle activity
recorded during teleoperation, without
requiring EMG at deployment.


\textbf{Human signals for robot learning.}
Human demonstrations provide information beyond
robot states and actions. DexCap~\cite{wang2024dexcapscalableportablemocap}
and DexWild~\cite{tao2025dexwild} leverage human
hand motion and interaction data for dexterous
manipulation, while
MimicTouch~\cite{yu2025mimictouchleveragingmultimodalhuman}
uses tactile demonstrations for contact-rich
tasks. Human gaze offers another complementary
signal: GazeVLA~\cite{li2026gazevlalearninghumanintention}
uses gaze to guide action prediction, while
GazeAct~\cite{pani2026gazeacthumangazeannotatedmanipulationdataset}
and Gaze-Regularized VLAs~\cite{pani2026gazeregularizedvisionlanguageactionmodelsrobotic}
explore gaze-derived supervision for visual
attention. In contrast, Robo-EMG investigates
the muscle activity underlying human
manipulation as an additional source of
supervision for VLA learning.


\textbf{EMG for robot control and learning.}
Surface EMG measures electrical activity
associated with muscle excitation and has
long supported movement recognition and
myoelectric control~\cite{204774,6737308}.
Its relationship with movement timing has
also motivated early movement prediction
~\cite{Norman1979,Smith2011,Koch2017}.
In robotics, \citet{beda2018} use EMG
commands for grasping, while
\citet{10.3389/fnins.2021.694914}
and \citet{ZHANG2022330} use EMG for
stiffness adaptation and interaction-force
estimation. Related demonstration-based
approaches leverage EMG to estimate human
stiffness, improve trajectory learning,
and transfer manipulation skills
~\cite{biagiotti,LIU2024127248,biomimetics10060399,Wang2025HumanRobotIS}.
More recently,
DexEMG~\cite{zhao2026dexemgdexterousteleoperationemg2pose}
predicts hand poses for dexterous
teleoperation, while
ForceBand~\cite{he2026forcebandlearningforcefulmanipulation}
estimates fingertip forces from EMG and
inertial measurements to enrich demonstrations
for robot policy learning. Unlike approaches
that primarily derive control signals,
stiffness, poses, or force estimates from
EMG, Robo-EMG uses temporally aligned
EMG representations to supervise VLA
multimodal features during training.

\begin{figure*}[t]
    \centering
    \includegraphics[width=\textwidth]{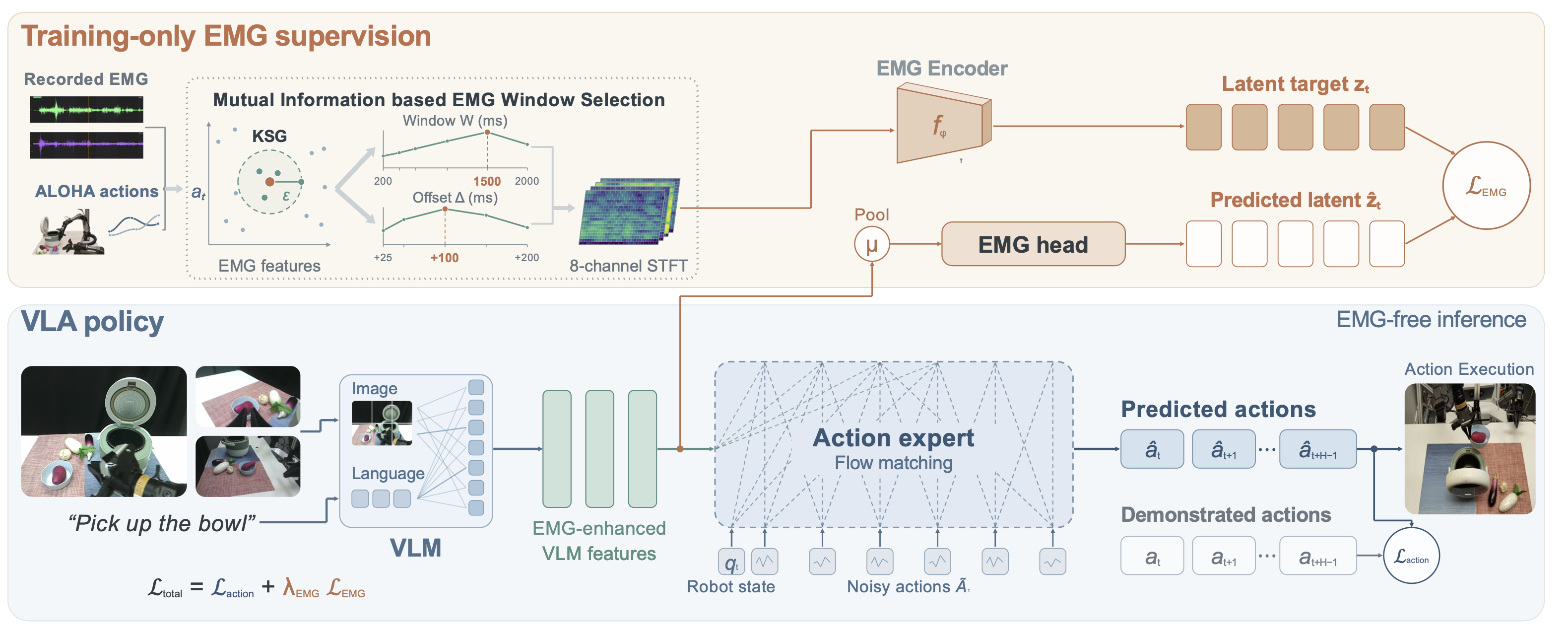}
\caption{
    \textbf{Robo-EMG training framework.}
    MI-guided temporal analysis identifies
    informative EMG windows from synchronized
    demonstrations, which a pretrained, frozen
    EMG encoder transforms into supervision
    targets. During VLA training, an auxiliary
    head predicts the aligned EMG embeddings
    from pooled VLM features, while the action
    expert learns to generate robot action chunks.
    The two objectives jointly guide policy
    learning. At inference, the auxiliary branch
    is removed, requiring no EMG measurements.
}
\vspace{-4mm}
    \label{fig:robo_emg_framework}
\end{figure*}

\textbf{Multimodal manipulation datasets.}
Large-scale robot datasets such as
Open X-Embodiment~\cite{embodimentcollaboration2025openxembodimentroboticlearning}
and DROID~\cite{khazatsky2025droidlargescaleinthewildrobot}
provide diverse robot observations and action
trajectories, but lack teleoperator EMG.
Other datasets capture human physiological
signals: Kaiwu~\cite{Jiang_2025_kaiwu}
records EMG, gaze, motion, and multimodal
observations during assembly tasks, while
MotorSense~\cite{dessalene2026motorsense}
pairs egocentric video with EMG for human
action understanding. Robo-EMG complements
these datasets by recording synchronized
human EMG alongside multi-view observations
and executable robot actions during real-world
and simulated teleoperation. This pairing
enables direct analysis of EMG-robot action
temporal relationships and provides aligned EMG
supervision targets for VLA training.

\section{Method}
\label{sec:method}

\methodname{} leverages human muscle activity recorded
during teleoperation to provide complementary
supervision for VLA policies. Rather than using
EMG as an additional input to the robot, we
investigate how muscle and physiological information can
guide representation learning during training.
As illustrated in Fig.~\ref{fig:robo_emg_framework},
our framework first identifies informative
temporal relationships between human EMG and
robot actions, then uses the aligned signals
to supervise the policy's multimodal features.

Our approach consists of four components.
We first collect synchronized robot demonstrations
and teleoperator EMG in real-world and simulated
environments (Sec.~\ref{sec:data_collection}). Since synchronization
alone may not establish which muscle activity
is most relevant to an action, we analyze
EMG-action dependence across window durations
and temporal offsets (Sec.~\ref{sec:temporal_dependence}). Based on the
selected alignments, we construct two alternative
supervision targets: sensor-wise EMG activation
and learned EMG embeddings (Sec.~\ref{sec:emg_representation}).
Finally, an auxiliary prediction head supervises
the VLM's representations alongside the original
action-learning objective (Sec.~\ref{sec:emg_vla}).
Importantly, EMG signals are used only during training,
allowing the policy to operate with its standard
inputs at inference.



\subsection{Synchronized Multimodal Data Collection}
\label{sec:data_collection}

We introduce Robo-EMG, a dataset that pairs
robot manipulation demonstrations with the
muscle activity of the human teleoperator, enabling us to study
the relationship between human muscle
activity and robot behavior.

\textbf{Real-world data collection.} 
For real-world collection, we use an ALOHA
leader-follower platform, where the operator
controls a single follower arm using its
corresponding leader arm. Each demonstration
records seven-dimensional robot actions. 10  teleoperators participate
in data collection, each providing 30
demonstrations per task across 15 tasks,
yielding 4,500 trajectories in total.
Each demonstration contains robot states,
action commands, and RGB observations from
three cameras: a central camera, a right
wrist-mounted camera, and a static left camera.

\textbf{Simulation data collection.}
We extend the same single-arm teleoperation procedure to MuJoCo,
where operators use the physical ALOHA
leader interface to control a simulated
robot while observing the environment
on a screen. Although the environment
is simulated, actions are human-generated
and EMG signals are recorded from operators,
rather than synthesized.
Ten teleoperators collected 1,000 demonstrations
across 20 manipulation tasks, which expands
to 30,000 trajectories through visual
augmentation. Further details on participant
allocation, data collection, and augmentation
are provided in Appendix~\ref{app:participants}.

\begin{figure}[t]
    \centering
    \includegraphics[width=\linewidth]{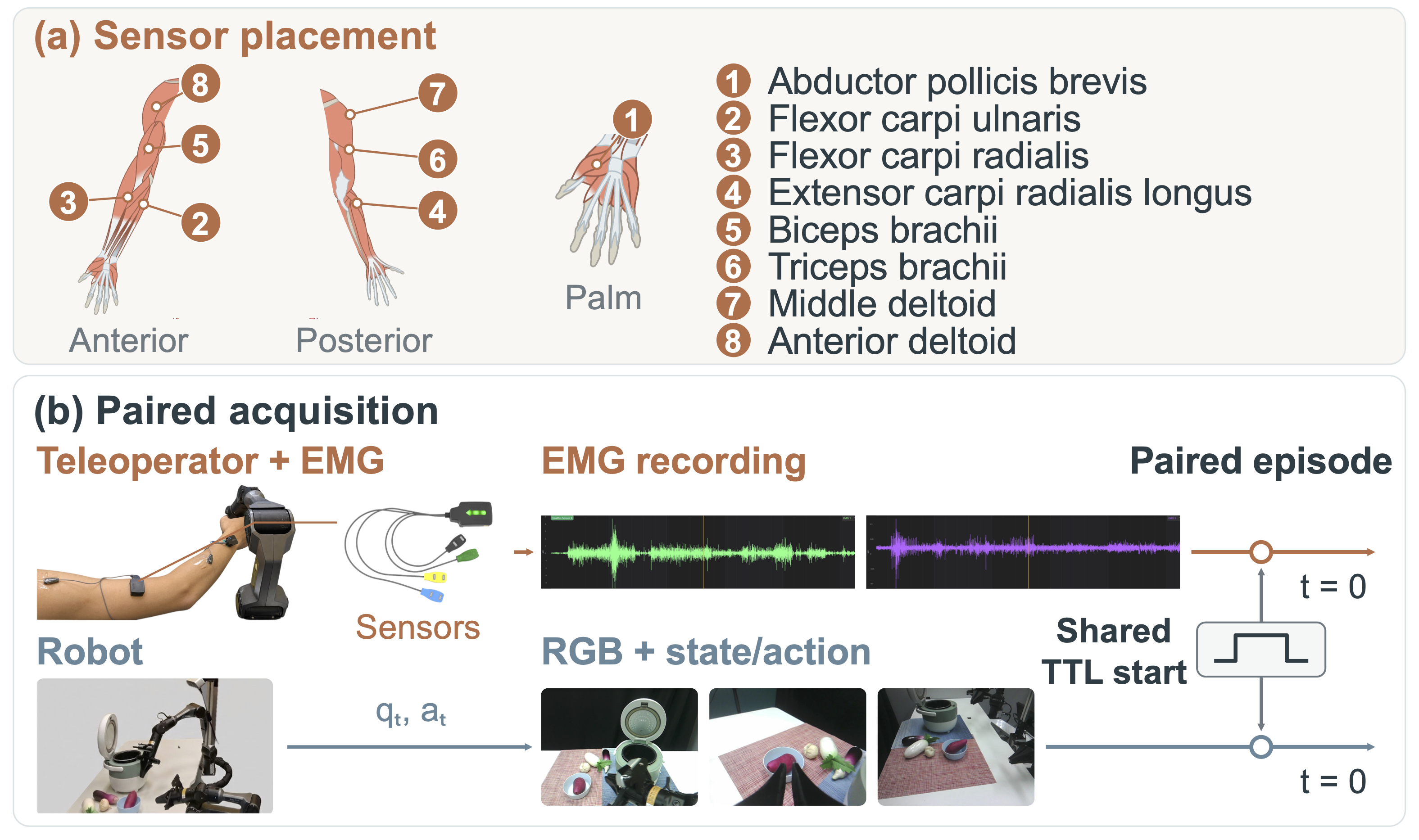}
\caption{
    \textbf{Robo-EMG muscle sensing and data collection.}
    (a) Eight upper-limb EMG recording sites cover
    muscles involved in thumb, wrist, elbow,
    and shoulder movements.
    (b) During ALOHA teleoperation, robot observations,
    states, and actions are recorded alongside
    eight-channel EMG at 500~Hz.
    TTL markers and timestamps align these streams,
    producing synchronized demonstrations for
    EMG-action analysis and policy supervision.
}
\vspace{-3mm}
    \label{fig:equipment-used}
\end{figure}

\textbf{EMG acquisition and synchronization.}
We record eight channels of surface EMG
at 500~Hz using a Delsys Trigno system.
Sensors are placed across the hand, forearm,
and upper arm to capture muscle activity
during grasping, object handling, and reaching.
For instance, the abductor pollicis brevis
contributes to thumb movement, while the
anterior deltoid supports arm positioning.
Additional sites capture wrist and elbow
activity. The eight recording sites are
illustrated in Fig.~\ref{fig:equipment-used},
following established surface-EMG practices
and reaching-to-grasp studies
\cite{HERMENS2000361,10.3389/fnins.2021.733359}.
For synchronization, hardware
transistor--transistor logic (TTL) markers
are recorded by the robot and EMG systems
at the beginning and end of each real-world
trial. Together with acquisition timestamps,
these markers align EMG and robot actions
on a common timeline for the temporal
analysis in Sec.~\ref{sec:temporal_dependence}.

\textbf{Ethics Compliance.} The study received prior institutional ethics
approval, and informed consent was obtained
from all participating teleoperators.
Further details of the ethics approval
will be disclosed upon acceptance to
preserve anonymity.

\subsection{EMG-Action Temporal Dependence}
\label{sec:temporal_dependence}

Although EMG and robot actions are synchronized
during collection using the hardware trigger, the muscle activity most
relevant to an action may occur earlier in time.
Prior studies have shown that muscle activation
can precede measurable mechanical responses
~\cite{Norman1979,Koirala2015}, while the
duration of EMG history influences movement
recognition and prediction
~\cite{Smith2011,Koch2017}. Motivated by these
observations, we investigate how EMG-action
dependence varies with window duration and
temporal offset to identify informative
alignments and optimal windows for robot policy supervision.

\textbf{Temporal windows and EMG activation.}
During each teleoperated demonstration, we
record the operator's EMG alongside the robot's
actions. Let $\mathbf{a}_t\in\mathbb{R}^{7}$
denote the robot action at time $t$.
Using synchronized timestamps, we associate
each action with a segment of the recorded
eight-channel EMG. To examine different
temporal relationships, we vary the duration
and offset of this segment:
\begin{equation}
    X_{W,\Delta}(t)
    = \mathrm{EMG}[t-\Delta-W:t-\Delta],
    \label{eq:emg_window}
\end{equation}
where $X_{W,\Delta}(t)\in
\mathbb{R}^{N_W\times 8}$ contains $N_W$
EMG samples, $W$ denotes the window duration,
and $\Delta$ denotes the offset between the
window's end and the action time. Varying
$W$ changes the amount of EMG history,
while $\Delta$ shifts the window relative
to the robot action.

We summarize each window using an
eight-dimensional EMG-activation vector:
\begin{equation}
    \mathbf{e}_t^{W,\Delta}
    = f_{\mathrm{act}}\bigl(
        X_{W,\Delta}(t)
    \bigr)
    \in \mathbb{R}^{8},
    \label{eq:emg_activation}
\end{equation}
where $f_{\mathrm{act}}$ is a fixed
feature-extraction function. It computes
the mean EMG envelope of each sensor within
the window and is then normalized. This provides a
consistent representation for comparing
different temporal configurations while making MI estimation more manageable.

\textbf{Mutual information and consensus.}
To identify which EMG windows are most informative
about robot actions, we evaluate a grid of
window durations and temporal offsets.
For each configuration $(W,\Delta)$, we measure
the mutual information (MI) between the
corresponding EMG activation and robot actions:
\begin{equation}
    I(W,\Delta)
    = I\left(
        \mathbf{e}_t^{W,\Delta};
        \mathbf{a}_t
    \right).
    \label{eq:emg_action_mi}
\end{equation}
We employ four complementary MI estimators:
KSG~\cite{Kraskov_2004}, MINE~\cite{MINE}, GCMI~\cite{GCMI1}, and SMILE~\cite{SMILE}. Since these
estimators differ in their assumptions and
numerical scales, we combine their relative
rankings rather than their raw estimates.
Specifically, we define a consensus score:
\begin{equation}
    S(W,\Delta)
    = \frac{1}{4}
    \sum_{k=1}^{4} r_k(W,\Delta),
    \label{eq:mi_consensus}
\end{equation}
where $r_k(W,\Delta)$ denotes the percentile
rank assigned by estimator $k$ across the
evaluated configurations.

Rather than relying on a single peak,
we identify a connected region around the
consensus maximum, retaining configurations
with $S(W,\Delta)\geq(1-\eta)S_{\max}$,
where $\eta$ controls the selection tolerance.
This identifies a set of nearby candidate
alignments with consistently strong
EMG-action dependence across estimators.
Within this region, we select a task-specific
alignment $(W^\star,\Delta^\star)$, using
held-out action-prediction probes as additional
evidence of the candidate windows' predictive
usefulness. The selected alignment determines
the EMG windows used to construct supervision
targets in Sec.~\ref{sec:emg_representation}.
These alignments are empirical choices rather
than direct measurements of physiological
delays. Further details on MI estimation,
predictive probes, and alignment selection
are provided in Appendix~\ref{app:emg_temporal}.

\subsection{Constructing EMG Supervision Targets}
\label{sec:emg_representation}

Having identified informative EMG-action
alignments, we next consider how to represent
the EMG signals for VLA
supervision. We explore two alternative targets:
sensor-wise EMG activation and learned EMG
embeddings. This allows us to examine whether
a simple summary of muscle activity is sufficient,
or whether learned representations provide
more useful information for policy learning.

\textbf{EMG Activation.}
Our first target directly summarizes the
activity recorded by each EMG sensor.
Using the feature-extraction function
$f_{\mathrm{act}}$ introduced in
Sec.~\ref{sec:temporal_dependence}, we obtain
\begin{equation}
    \mathbf{y}_t^{\mathrm{act}}
    =
    f_{\mathrm{act}}\left(
        X_{W^\star,\Delta^\star}(t)
    \right)
    =
    \mathbf{e}_t^{W^\star,\Delta^\star}
    \in \mathbb{R}^{8},
    \label{eq:emg_activation_target}
\end{equation}
where $(W^\star,\Delta^\star)$ denotes the
selected temporal alignment. Each dimension
represents the normalized mean EMG envelope
of one sensor, preserving the correspondence
between the representation and the recorded
muscle activity.

\textbf{Learned EMG Embedding.}
While sensor-wise activation summarizes muscle
activity, averaging the EMG envelope can discard
finer temporal and spectral patterns.
Recent studies have demonstrated the value of
self-supervised representation learning for
capturing EMG dynamics without relying on
extensive manual annotations
~\cite{RAGHU2025109479,Yang2025-eh}.
Motivated by these findings and advances in
masked spectrogram modeling
~\cite{NEURIPS2022_b89d5e20,he2026forcebandlearningforcefulmanipulation}, we train a masked
spectrogram autoencoder to learn richer EMG
representations directly from recorded
muscle activity, without robot-action labels.

We transform eight-channel EMG windows into
log-power spectrograms using the short-time
Fourier transform (STFT). A Transformer-based
encoder learns representations through masked
spectrogram reconstruction. The resulting
features are projected into a compact
EMG embedding.
After pretraining, we freeze the EMG encoder
$f_\phi$ and use it to construct the
supervision target from the selected window:
\begin{equation}
    \mathbf{z}_t^{\mathrm{EMG}}
    =
    f_\phi\left(
        X_{W^\star,\Delta^\star}(t)
    \right)
    \in \mathbb{R}^{d_z},
    \label{eq:emg_embedding}
\end{equation}

where $d_z=128$ denotes the learned embedding
dimension. The window duration $W^\star$ determines the
signal history processed by the encoder,
while $\Delta^\star$ determines its alignment
with the corresponding robot action.
Both representations are used as alternative
auxiliary targets during VLA training, as
described in Sec.~\ref{sec:emg_vla}.
Further details on target construction
and encoder pretraining are provided
in the Appendix Sec.~\ref{app:emg_encoder}.

\subsection{EMG Supervision for VLA Policies}
\label{sec:emg_vla}

Having constructed temporally aligned EMG
representations, we use them as auxiliary
supervision targets during VLA training.
Our goal is to encourage the policy's visual
and language representations to capture
information associated with the demonstrator's
muscle activity, while preserving the original
action-generation architecture. The overall
training framework is illustrated in
Fig.~\ref{fig:robo_emg_framework}.

\textbf{Auxiliary EMG prediction.}
At each timestep $t$, the VLA processes the
current visual observations and language
instruction to predict an action chunk
$\mathbf{A}_t =
[\mathbf{a}_t,\ldots,\mathbf{a}_{t+H-1}]
\in \mathbb{R}^{H\times 7}$
of length $H$. The VLM's multimodal features
condition the action expert in generating
these actions. We therefore investigate
whether the same features can also capture
information associated with the demonstrator's
muscle activity.

To encourage this, we introduce an auxiliary
task that predicts the EMG representation
aligned with the starting action
$\mathbf{a}_t$. Rather than supervising
the action expert directly, we attach the
prediction head to the VLM, allowing the
additional learning signal to influence
the features used for action generation.
Specifically, we apply pooling
to the VLM's final-layer image and language
token features to obtain a representation
$\mathbf{h}_t$. An auxiliary EMG head then
predicts the aligned EMG target:
\begin{equation}
    \hat{\mathbf{y}}_t^{\mathrm{EMG}}
    = f_{\mathrm{aux}}(\mathbf{h}_t),
    \label{eq:emg_prediction}
\end{equation}
where $\mathbf{y}_t^{\mathrm{EMG}}$
denotes either the EMG-activation target
$\mathbf{y}_t^{\mathrm{act}}$ or the
learned EMG embedding
$\mathbf{z}_t^{\mathrm{EMG}}$
from Sec.~\ref{sec:emg_representation}.
Both targets are constructed from EMG windows
temporally aligned with the starting action
$\mathbf{a}_t$ and are used in separate
training configurations.

\textbf{Joint training objective.}
We jointly optimize the original
action-learning objective and the auxiliary
EMG prediction loss:
\begin{equation}
    \mathcal{L}_{\mathrm{total}}
    =
    \mathcal{L}_{\mathrm{action}}
    +
    \lambda_{\mathrm{EMG}}
    \mathcal{L}_{\mathrm{EMG}},
    \label{eq:joint_training}
\end{equation}
where $\mathcal{L}_{\mathrm{EMG}}$
is the mean-squared error between the
predicted and target EMG representations,
and $\lambda_{\mathrm{EMG}}$ controls
the strength of auxiliary supervision.

\begin{table*}[t]
    \centering
    \caption{
        Comparison of Robo-EMG with physiological
        and robot manipulation datasets. Recording
        time is reported in hours where available;
        dataset scale retains each source's native
        unit. Paired EMG-action indicates synchronized
        human EMG and robot actions.
    }
    \label{tab:dataset_comparison}

    \footnotesize
    \setlength{\tabcolsep}{2.8pt}
    \renewcommand{\arraystretch}{1.28}

    \begin{tabularx}{\textwidth}{
        @{}
        >{\raggedright\arraybackslash}p{0.168\textwidth}
        >{\centering\arraybackslash}p{0.079\textwidth}
        >{\raggedright\arraybackslash}p{0.090\textwidth}
        *{4}{>{\centering\arraybackslash}p{0.059\textwidth}}
        >{\centering\arraybackslash}p{0.062\textwidth}
        >{\centering\arraybackslash}p{0.076\textwidth}
        >{\raggedright\arraybackslash}X
        @{}
    }
        \toprule

        \multirow{2}{*}{\textbf{Dataset}} &
        \multirow{2}{*}{\textbf{Setting}} &
        \multirow{2}{*}{\textbf{Robot}} &
        \multicolumn{4}{c}{\textbf{Available Modalities}} &
        \multirow{2}{*}{
            \shortstack{\textbf{EMG}\\\textbf{(Hz)}}
        } &
        \multirow{2}{*}{
            \shortstack{\textbf{Time}\\\textbf{(h)}}
        } &
        \multirow{2}{*}{
            \shortstack[l]{
                \textbf{Dataset Scale}\\
                \textbf{(native units)}
            }
        } \\

        \cmidrule(lr){4-7}

        & & &
        \shortstack{\textbf{Human}\\\textbf{EMG}} &
        \shortstack{\textbf{Robot}\\\textbf{actions}} &
        \shortstack{\textbf{Multi-}\\\textbf{view}\\\textbf{RGB}} &
        \shortstack{\textbf{Paired}\\\textbf{EMG--}\\\textbf{action}} &
        & & \\

        \midrule

        Kaiwu ~\cite{Jiang_2025_kaiwu}
            & Real
            & --
            & $\checkmark$
            & $\times$
            & $\checkmark$
            & $\times$
            & 40
            &6.3
            & \shortstack[l]{
            300 Trajectories
            } \\

        \rowcolor{RoboGray}
        MotorSense ~\cite{dessalene2026motorsense}
            & Real
            & --
            & $\checkmark$
            & $\times$
            & $\times$
            & $\times$
            & 500
            & 70.4
            & 370 trajectories (public) \\

        ForceBand ~\cite{he2026forcebandlearningforcefulmanipulation}
            & Real
            & --
            & $\checkmark$
            & $\times$
            & $\times$
            & $\times$
            & 250
            & 10
            & 135 trajectories \\

        \rowcolor{RoboGray}
        ALOHA / ACT ~\cite{zhao2023learningfinegrainedbimanualmanipulation}
            & Real + Sim
            & ALOHA
            & $\times$
            & $\checkmark$
            & $\checkmark$
            & $\times$
            & --
            & n/a
            & \shortstack[l]{
               825 Trajectories
            } \\

        DROID ~\cite{khazatsky2025droidlargescaleinthewildrobot}
            & Real
            & Franka
            & $\times$
            & $\checkmark$
            & $\checkmark$
            & $\times$
            & --
            & 350
            & 76,000 trajectories \\

        \rowcolor{RoboGray}
        Open X-Embodiment ~\cite{embodimentcollaboration2025openxembodimentroboticlearning}
            & Real
            & Multiple
            & $\times$
            & $\checkmark$
            & Varies
            & $\times$
            & --
            & n/a
            & 1,000,000+ trajectories \\

        \midrule

        \rowcolor{RoboBlue}
        \textbf{Robo-EMG (Ours)}
            & \textbf{Real + Sim}
            & \shortstack[l]{
                \textbf{ALOHA /}\\[-1pt]
                \textbf{Franka}
            }
            & $\checkmark$
            & $\checkmark$
            & $\checkmark$
            & $\checkmark$
            & \textbf{500}
            &  55
            & \shortstack[l]{
                \textbf{5,500 trajectories}\\[-1pt]
                {\scriptsize 4,500 real + 1,000 sim}
            } \\

        \bottomrule
    \end{tabularx}


\end{table*}

The auxiliary loss updates the prediction
head and trainable VLM parameters,
encouraging the shared multimodal features
to retain information associated with
the demonstrator's muscle activity.
Because EMG serves only as a training
target, the auxiliary branch is removed
at inference, leaving the standard VLA
action-generation pathway unchanged.
Further implementation details are
provided in the Appendix in Sec.~\ref{app:vla_training}.

\section{Experiments}
\label{sec:experiments}

\textbf{Policy training.}
We use $\pi_{0.5}$ as the primary VLA backbone
for both real-world and simulation experiments.
To evaluate the effect of EMG supervision,
we compare a baseline policy trained with the
original action-learning objective against
an EMG-supervised variant that additionally
predicts temporally aligned EMG representations,
as described in Sec.~\ref{sec:emg_vla}.
Both policies use the same visual, language,
and robot-state inputs at inference, without
requiring EMG measurements.
We further evaluate whether the benefits extend
across different VLA architectures.

\textbf{Real-world evaluation.}
We evaluate manipulation performance on our
ALOHA platform using strict task success
rates, defined as the percentage of rollouts
that satisfy task-specific completion criteria.
Our primary comparison covers nine manipulation
tasks, with 50 evaluation rollouts per task.
We also conduct controlled studies on task
subsets to examine the effects of EMG
supervision strength and temporal alignment, robustness to camera-view occlusion and generalization
across VLA backbones.
For the occlusion study, we evaluate
10 trials per task and visual condition.

\textbf{Simulation evaluation.}
We train baseline and EMG-supervised policies
using our human-teleoperated MuJoCo dataset,
with recorded EMG providing auxiliary
supervision for the latter. We then evaluate
both policies through robot rollouts under
clean and visually perturbed conditions,
measuring task success on held-out simulation
tasks and the LIBERO Spatial benchmark.
Success is determined by task completion
during execution, rather than predictions
on recorded demonstrations. Further details
on the evaluation protocols, the simulation setup and visual
perturbations are provided in
Appendix~\ref{app:simulation_protocol}.

\subsection{Dataset Comparison}
\label{sec:dataset_comparison}

As shown in Table~\ref{tab:dataset_comparison},
existing physiological datasets capture human
muscle activity but generally lack corresponding
robot actions, while robot manipulation datasets
lack synchronized teleoperator EMG.
Robo-EMG bridges this gap by pairing
eight-channel human EMG with multi-view
observations and robot actions.
It contains 5,500 human-teleoperated
demonstrations across 35 tasks, including
4,500 real-world and 1,000 simulated
trajectories. Importantly, EMG in both
settings is recorded from human operators,
enabling temporal dependence analysis and
training-only physiological supervision.

\subsection{Real-World Manipulation Experiments}
\label{sec:real_world}

\textbf{Effectiveness of EMG Supervision.}
\label{sec:real_emg_effectiveness}
We first investigate whether human muscle
activity can improve manipulation when used
solely as training supervision. We compare
three policies sharing the same VLA backbone:
Baseline, trained without EMG; EMG Activation,
using sensor-wise activation targets; and
EMG Encoder, using learned EMG embeddings.
All models are evaluated on nine real-world
tasks, with 50 rollouts per task.
As shown in Table~\ref{tab:real_world_comparison},
both EMG-supervised variants outperform the
baseline (40.9\%), achieving 47.1\% with
EMG Activation and 58.9\% with EMG Encoder.
Notably, the learned representation performs
best on all nine tasks, improving average
success by 18.0 percentage points.
This suggests that learned EMG representations
provide a more effective auxiliary signal
than sensor-wise activation alone.

\begin{table*}[t]
    \centering
    \caption{
        Real-world manipulation success rates (\%)
        across nine tasks, evaluated over 50 rollouts
        per task. Values in brackets are 95\% Wilson
        confidence intervals. $W$ and $\Delta$ are
        the task-specific EMG Encoder alignment
        parameters (ms).
    }
    \label{tab:real_world_comparison}

    
        \small
        \setlength{\tabcolsep}{3.5pt}
        \renewcommand{\arraystretch}{1.12}
        
        \renewcommand{\srci}[2]{%
            \mbox{#1\hspace{2pt}{\scriptsize [#2]}}%
        }

    \begin{tabularx}{\textwidth}{
        @{}
        >{\raggedright\arraybackslash}p{0.32\textwidth}
        >{\centering\arraybackslash}X
        >{\centering\arraybackslash}X
        >{\columncolor{RoboBlue}\centering\arraybackslash}X
        >{\centering\arraybackslash}p{0.065\textwidth}
        >{\centering\arraybackslash}p{0.065\textwidth}
        @{}
    }
        \toprule

        \multirow{2}{*}{\textbf{Manipulation Task}} &
        \multicolumn{3}{c}{\textbf{Task Success Rate (\%)}} &
        \multicolumn{2}{c}{\textbf{Encoder Alignment (ms)}} \\

        \cmidrule(lr){2-4}
        \cmidrule(l){5-6}

        & \textbf{Baseline}
        & \textbf{EMG Activation}
        & \textbf{EMG Encoder}
        & \textbf{$W$}
        & \textbf{$\Delta$} \\

        \midrule

        \tasklabel{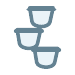}{Stack three cups}
        & \srci{48}{34.8, 61.5}
        & \srci{56}{42.3, 68.8}
        & \srci{\textbf{70}}{56.2, 80.9}
        & 1500 & 50 \\

        \tasklabel{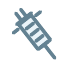}{Place syringe on hand}
        & \srci{46}{33.0, 59.6}
        & \srci{64}{50.1, 75.9}
        & \srci{\textbf{88}}{76.2, 94.4}
        & 1500 & 50 \\

        \tasklabel{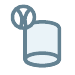}{Place tennis ball in can}
        & \srci{32}{20.8, 45.8}
        & \srci{36}{24.1, 49.9}
        & \srci{\textbf{48}}{34.8, 61.5}
        & 1500 & 0 \\

        \tasklabel{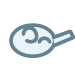}{Place food item on pan}
        & \srci{50}{36.6, 63.4}
        & \srci{54}{40.4, 67.0}
        & \srci{\textbf{60}}{46.2, 72.4}
        & 1500 & 100 \\

        \tasklabel{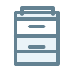}{Open top drawer}
        & \srci{54}{40.4, 67.0}
        & \srci{54}{40.4, 67.0}
        & \srci{\textbf{66}}{52.2, 77.6}
        & 1500 & 0 \\

        \tasklabel{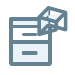}{Open drawer and place cube}
        & \srci{44}{31.2, 57.7}
        & \srci{48}{34.8, 61.5}
        & \srci{\textbf{56}}{42.3, 68.8}
        & 1000 & 100 \\

        \tasklabel{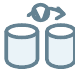}{Transfer tennis ball between containers}
        & \srci{16}{8.3, 28.5}
        & \srci{16}{8.3, 28.5}
        & \srci{\textbf{20}}{11.2, 33.0}
        & 1500 & 0 \\

        \tasklabel{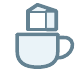}{Place sugar cube in cup}
        & \srci{58}{44.2, 70.6}
        & \srci{64}{50.1, 75.9}
        & \srci{\textbf{80}}{67.0, 88.8}
        & 1000 & 100 \\

        \tasklabel{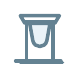}{Place cup in cup stand}
        & \srci{20}{11.2, 33.0}
        & \srci{32}{20.8, 45.8}
        & \srci{\textbf{42}}{29.4, 55.8}
        & 500 & 50 \\

        \midrule

        \rowcolor{RoboGray}
        \textbf{Mean}
        & \textbf{40.9}
        & \textbf{47.1}
        & \cellcolor{RoboBlue}\textbf{58.9}
        & -- & -- \\

        \bottomrule
    \end{tabularx}
\end{table*}


\textbf{Effect of Auxiliary Loss Weight.}
\label{sec:real_loss_weight}
We investigate how the strength of EMG
supervision affects manipulation performance
by varying the auxiliary loss weight
$\lambda_{\mathrm{EMG}} \in
\{0, 0.01, 0.1, 10\}$, where
$\lambda_{\mathrm{EMG}}=0$ corresponds
to the baseline without EMG supervision.
As shown in Table~\ref{tab:real_loss_weight},
$\lambda_{\mathrm{EMG}}=0.1$ achieves the
highest mean success rate of 68.0\% across
five real-world tasks, compared with 48.4\%
for the baseline. Lower and higher weights
yield 48.0\% and 54.0\%, respectively.
These results suggest that EMG supervision
is most beneficial when appropriately
balanced with the primary action-learning
objective. Excessive weighting may shift
optimization away from action prediction.
Further training details are provided in
the supplementary material.

\begin{table}[t]
    \centering
    \caption{
        Effect of auxiliary EMG loss weight on
        real-world manipulation performance.
        Values indicate task success rates
        (\%).
    }
    \label{tab:real_loss_weight}

    \footnotesize
    \setlength{\tabcolsep}{3pt}
    \renewcommand{\arraystretch}{1.28}

    \begin{tabularx}{\columnwidth}{
        @{}
        >{\raggedright\arraybackslash}X
        >{\centering\arraybackslash}p{0.115\columnwidth}
        >{\centering\arraybackslash}p{0.115\columnwidth}
        >{\columncolor{RoboBlue}\centering\arraybackslash}p{0.135\columnwidth}
        >{\centering\arraybackslash}p{0.115\columnwidth}
        @{}
    }
        \toprule

        \multirow{2}{*}{\textbf{Manipulation Task}} &
        \multicolumn{4}{c}{
            \textbf{EMG Loss Strength}
            ($\lambda_{\mathrm{EMG}}$)
        } \\

        \cmidrule(l){2-5}

        & $\mathbf{0}$
        & $\mathbf{0.01}$
        & $\mathbf{0.1}$
        & $\mathbf{10}$ \\

        \midrule

        Stack three cups
            & 48 & 44 & \textbf{70} & 54 \\

        Place syringe on hand
            & 46 & 46 & \textbf{88} & 62 \\

        Place food item on pan
            & 50 & 52 & \textbf{60} & 48 \\

        Open top drawer
            & 54 & 52 & \textbf{66} & 58 \\

        Open drawer and place cube
            & 44 & 46 & \textbf{56} & 48 \\

        \midrule

        \rowcolor{RoboGray}
        \textbf{Mean}
            & \textbf{48.4}
            & \textbf{48.0}
            & \cellcolor{RoboBlue}\textbf{68.0}
            & \textbf{54.0} \\

        \bottomrule

    \end{tabularx}
\end{table}





\textbf{Effect of EMG Temporal Alignment.}
\label{sec:real_temporal_alignment}
We next investigate whether selecting
task-specific EMG windows improves policy
performance. We compare alignments identified
through MI consensus with two fixed settings:
$(W,\Delta)=(200,0)$ and $(400,0)$ ms.
As shown in Table~\ref{tab:real_temporal_alignment},
the selected alignments achieve the highest
success rates across all five tasks, reaching
68.0\% on average compared with 58.4\%
and 53.6\% for the fixed settings.
These results suggest that
task-dependent EMG-action timing provides
more effective supervision.

\begin{table}[t]
    \centering
    \caption{
        Effect of EMG temporal alignment on real-world
        manipulation success (\%). We compare fixed
        configurations with the task-specific alignment
        selected using MI consensus.
    }
    \label{tab:real_temporal_alignment}

    \footnotesize
    \setlength{\tabcolsep}{3pt}
    \renewcommand{\arraystretch}{1.28}

    \begin{tabularx}{\columnwidth}{
        @{}
        >{\raggedright\arraybackslash}X
        >{\centering\arraybackslash}p{0.17\columnwidth}
        >{\centering\arraybackslash}p{0.17\columnwidth}
        >{\columncolor{RoboBlue}\centering\arraybackslash}
            p{0.19\columnwidth}
        @{}
    }
        \toprule

        \textbf{Manipulation Task} &
        \multicolumn{3}{c}{
            \textbf{Encoder Alignment $(W,\Delta)$ (ms)}
        } \\

        \cmidrule(l){2-4}

        &
        $\mathbf{(200,0)}$ &
        $\mathbf{(400,0)}$ &
        \textbf{Selected} \\

        \midrule

        Stack three cups
            & 62 & 50 & \textbf{70} \\

        Place syringe on hand
            & 64 & 56 & \textbf{88} \\

        Place food item on pan
            & 54 & 54 & \textbf{60} \\

        Open top drawer
            & 62 & 60 & \textbf{66} \\

        Open drawer and place cube
            & 50 & 48 & \textbf{56} \\

        \midrule

        \rowcolor{RoboGray}
        \textbf{Mean}
            & \textbf{58.4}
            & \textbf{53.6}
            & \cellcolor{RoboBlue}\textbf{68.0} \\

        \bottomrule
    \end{tabularx}
\end{table}




\textbf{Robustness to Camera View Occlusion.}
\label{sec:real_robustness}
We next examine whether EMG supervision
improves robustness when visual information
is partially unavailable. We evaluate baseline
and EMG-supervised policies on four real-world
tasks under clean observations and with one of the
camera views occluded, using 10 trials per
task.
As shown in Table~\ref{tab:real_robustness},
EMG supervision improves mean success under
occlusion from 37.5\% to 72.5\%.
More importantly, occlusion reduces baseline
success by 25.0 percentage points, compared
with only 10.0 points for the EMG-supervised
policy. These results suggest that EMG-based
training helps the policy remain effective
when visual information is incomplete,
despite requiring no EMG at inference. 
Further details on the camera-occlusion
conditions and evaluation protocol are
provided in Appendix~\ref{app:camera_occlusion}.

\begin{table}[t]
    \centering
    \caption{
Real-world manipulation success rates (\%)
under clean observations and single-camera
occlusion across four manipulation tasks.
Each condition is evaluated over 10 trials
per task using identical success criteria.
    }
    \label{tab:real_robustness}

    \footnotesize
    \setlength{\tabcolsep}{3pt}
    \renewcommand{\arraystretch}{1.28}

    \begin{tabularx}{\columnwidth}{
        @{}
        >{\raggedright\arraybackslash}X
        >{\centering\arraybackslash}
            p{0.13\columnwidth}
        >{\columncolor{RoboBlue}\centering\arraybackslash}
            p{0.13\columnwidth}
        >{\centering\arraybackslash}
            p{0.13\columnwidth}
        >{\columncolor{RoboBlue}\centering\arraybackslash}
            p{0.13\columnwidth}
        @{}
    }
        \toprule

        \multirow{2}{*}{\textbf{Manipulation Task}} &
        \multicolumn{2}{c}{\textbf{Clean}} &
        \multicolumn{2}{c}{\textbf{Occluded}} \\

        \cmidrule(lr){2-3}
        \cmidrule(l){4-5}

        &
        \textbf{Base} &
        \textbf{EMG} &
        \textbf{Base} &
        \textbf{EMG} \\

        \midrule

        Stack three cups
            & 80 & \textbf{90}
            & 40 & \textbf{80} \\

        Place syringe on hand
            & 60 & \textbf{90}
            & 30 & \textbf{80} \\

        Open top drawer
            & 50 & \textbf{70}
            & 50 & \textbf{60} \\

        Open drawer place cube
            & 60 & \textbf{80}
            & 30 & \textbf{70} \\

        \midrule

        \rowcolor{RoboGray}
        \textbf{Mean}
            & \textbf{62.5}
            & \cellcolor{RoboBlue}\textbf{82.5}
            & \textbf{37.5}
            & \cellcolor{RoboBlue}\textbf{72.5} \\

        \bottomrule
    \end{tabularx}
    \vspace{-4mm}
\end{table}



\textbf{Generalization Across VLAs.}
\label{sec:cross_backbone}
To examine whether the benefits of EMG
supervision extend beyond $\pi_{0.5}$,
we evaluate two other VLA backbones:
$\pi_0$ and OpenVLA-OFT.
For each backbone, we compare its baseline
policy with an EMG-supervised variant
on real-world manipulation tasks.
As shown in Table~\ref{tab:cross_backbone},
EMG supervision improves success across
all three tasks and backbones. Mean success
increases from 42.7\% to 62.0\% for
$\pi_0$, 49.3\% to 74.7\% for $\pi_{0.5}$,
and 33.3\% to 48.7\% for OpenVLA-OFT.
These consistent gains suggest that EMG
provides a useful auxiliary training signal
across different VLA architectures.

\begin{table}[t]
    \centering
\caption{
    Cross-backbone comparison on 
    real-world tasks. Success rates (\%)
    are reported for each baseline (Base)
    and its EMG embedding variant (Ours).
    Cups: stack three cups; Syringe: place
    syringe on hand; Drawer: open top drawer.
}
    \label{tab:cross_backbone}

    \footnotesize
    \setlength{\tabcolsep}{3pt}
    \renewcommand{\arraystretch}{1.16}

    \begin{tabularx}{\columnwidth}{
        @{}
        >{\raggedright\arraybackslash}X
        cccc
        @{}
    }
        \toprule
        \textbf{Model} &
        \textbf{Cups} &
        \textbf{Syringe} &
        \textbf{Drawer} &
        \textbf{Mean} \\
        \midrule

        $\pi_0$ (Base)~\cite{black2026pi0visionlanguageactionflowmodel}
            & 42 & 38 & 48 & 42.7 \\
        \rowcolor{RoboBlue}
        $\pi_0$ (Ours)
            & 58 & 70 & 58 & 62.0 \\

        \addlinespace[3pt]

        $\pi_{0.5}$ (Base) ~\cite{intelligence2025pi05visionlanguageactionmodelopenworld}
            & 48 & 46 & 54 & 49.3 \\
        \rowcolor{RoboBlue}
        $\pi_{0.5}$ (Ours)
            & 70 & 88 & 66 & 74.7 \\

        \addlinespace[3pt]

        OpenVLA-OFT (Base) ~\cite{kim2025finetuningvisionlanguageactionmodelsoptimizing}
            & 34 & 28 & 38 & 33.3 \\
        \rowcolor{RoboBlue}
        OpenVLA-OFT (Ours)
            & 46 & 52 & 48 & 48.7 \\

        \bottomrule
    \end{tabularx}
\end{table}

\begin{figure}[t]
    \centering
    \includegraphics[
        width=\columnwidth,
        height=0.25\textheight,
        keepaspectratio
    ]{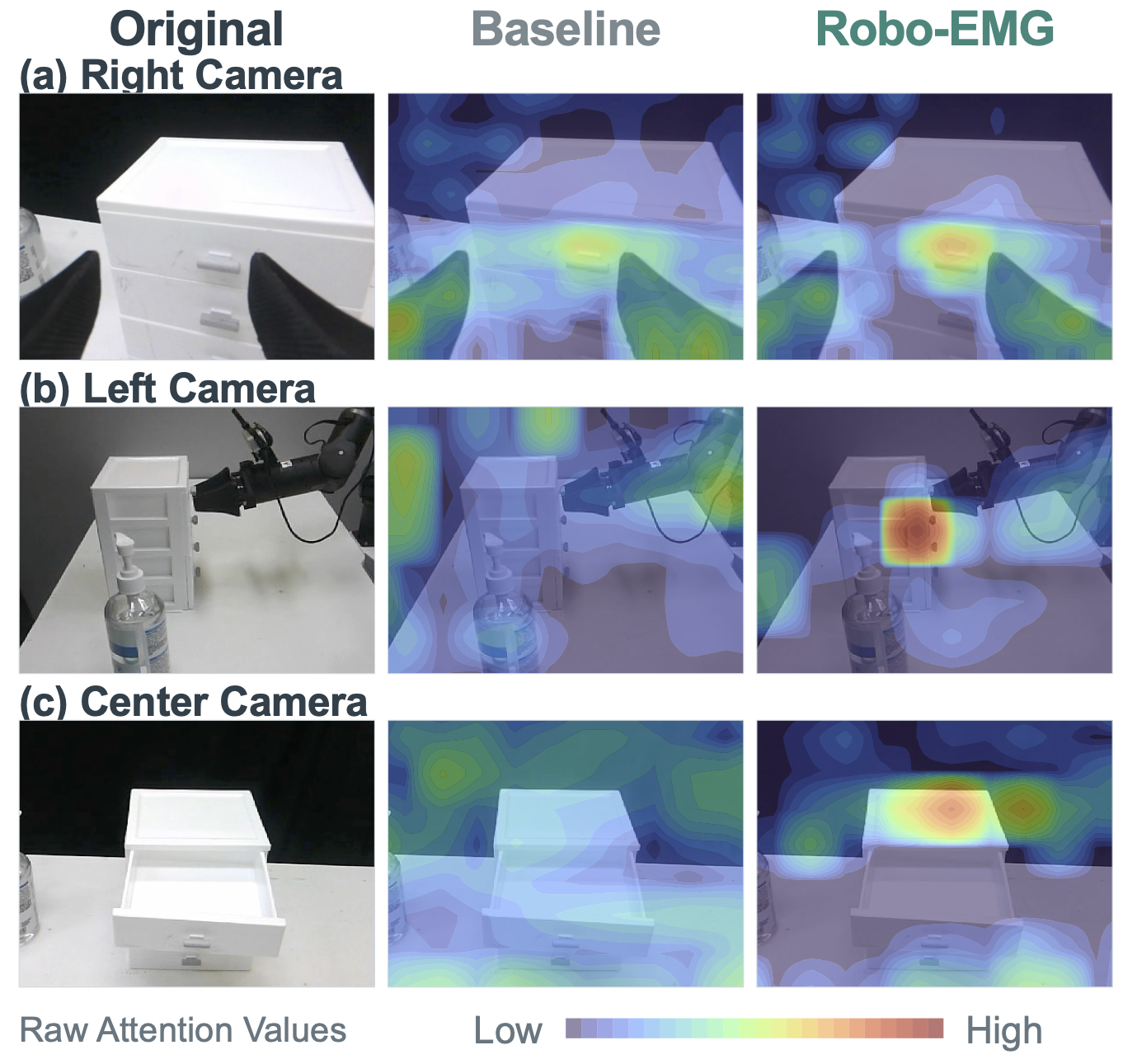}
    \caption{
        \textbf{Visual attention during drawer opening.}
        RGB observations and attention maps from
        Baseline and Robo-EMG across three camera
        views. Robo-EMG exhibits more localized
        responses around drawer-related regions.
    }
    \vspace{-4mm}
    \label{fig:attention_analysis}
\end{figure}

\textbf{Qualitative Attention Analysis.}
\label{sec:attention_analysis}
We also investigate how EMG supervision influences
the VLM's visual attention. As shown in
Fig.~\ref{fig:attention_analysis}, Robo-EMG
exhibits more localized responses around
drawer-related regions, particularly in the
left and center camera views, while baseline
attention is more dispersed across the scene.
Across sampled observations, Robo-EMG
also exhibits more concentrated attention,
while the baseline shows more diffuse
attention on average, as reflected by their
normalized spatial attention entropy
(0.412 vs. 0.753, respectively).
These observations suggest that EMG supervision
encourages more spatially focused visual
representations. Further attention analysis is provided
in Appendix~\ref{app:attention_analysis}.

\subsection{Simulation Experiments}
\label{sec:simulation_experiments}

We further evaluate \methodname{} in simulated
environments using human-teleoperated demonstrations
paired with recorded EMG. We train the policies on our human-teleoperated
simulation dataset and evaluate them under
both clean and visually perturbed conditions.
We measure task success through policy
rollouts on held-out simulation tasks
and the LIBERO Spatial benchmark.
As shown in Table~\ref{tab:simulation_results},
EMG supervision improves performance across both
evaluations and visual conditions. Under clean
conditions, held-out task success increases
from 69.7\% to 85.6\%, while LIBERO Spatial
success rises from 52.0\% to 60.0\%.
These gains persist under visual perturbations,
with improvements of 17.6 and 13.6 percentage
points, respectively. Together, these results
suggest that EMG supervision provides a useful
training signal across simulated manipulation
tasks and varying visual conditions.
Further details on the simulation evaluation protocols are provided in
Appendix~\ref{app:simulation_protocol}.

\begin{table}[t]
    \centering
\caption{
    Simulation task success rates (\%) under
    clean and visually perturbed evaluation settings. Baseline and
    Robo-EMG are compared
    on held-out tasks and LIBERO Spatial.
}
    \label{tab:simulation_results}

    \footnotesize
    \setlength{\tabcolsep}{3.5pt}
    \renewcommand{\arraystretch}{1.28}

    \begin{tabularx}{\columnwidth}{
        @{}
        >{\raggedright\arraybackslash}p{0.19\columnwidth}
        >{\raggedright\arraybackslash}X
        >{\centering\arraybackslash}p{0.19\columnwidth}
        >{\centering\arraybackslash}p{0.17\columnwidth}
        @{}
    }
        \toprule

        \textbf{Setting} &
        \textbf{Model} &
        \shortstack{\textbf{Held-out}\\\textbf{SR (\%)}} &
        \shortstack{\textbf{LIBERO}\\\textbf{SR (\%)}} \\

        \midrule

        \multirow{2}{*}{Clean}
            & Baseline
            & 69.7
            & 52.0 \\

            & \cellcolor{RoboBlue}\textbf{Ours (EMG)}
            & \cellcolor{RoboBlue}\textbf{85.6}
            & \cellcolor{RoboBlue}\textbf{60.0} \\

        \midrule

        \multirow{2}{*}{Perturbed}
            & Baseline
            & 55.8
            & 40.4 \\

            & \cellcolor{RoboBlue}\textbf{Ours (EMG)}
            & \cellcolor{RoboBlue}\textbf{73.4}
            & \cellcolor{RoboBlue}\textbf{54.0} \\

        \bottomrule
    \end{tabularx}
    \vspace{-4mm}
\end{table}

\section{Conclusion}
\label{sec:conclusion}

We introduce Robo-EMG, a dataset of 5,500
human-teleoperated demonstrations paired with
synchronized muscle activity across real-world
and simulated manipulation environments.
Building on this dataset, we characterize
informative EMG-action temporal relationships
and develop a training-only supervision
framework using aligned EMG representations
to enhance VLA learning. By supervising the
VLM's multimodal features with human muscle
activity, our approach improves policy learning
without requiring EMG at inference.
Across nine real-world tasks, learned EMG
supervision increases mean success from
40.9\% to 58.9\%, with additional gains
under camera occlusion, across VLA backbones,
and in simulated evaluations. These findings
highlight the potential of human physiological
signals to enrich robot learning beyond
conventional demonstrations. We hope this
work inspires further exploration of human
motor information for more capable and
adaptable manipulation policies.
We will publicly release the Robo-EMG dataset
and code to support future research and
reproducibility.

\section{LLM Usage}
Generative AI and LLM was used for helping to write the paper. ChatGPT was used to assist with making the writing easier to interpret, with some basic restructuring and for improving the clarity and presentation of the overall look of the paper. Generative AI tools were also used to create pencil-style sketches incorporated into Figure 2. All AI-assisted content was reviewed and edited by the authors. The research methodology, data collection, experimental execution, results, and scientific conclusions remains our responsibility.
{
    \small
    \bibliographystyle{ieeenat_fullname}
    \bibliography{main}

@misc{black2026pi0visionlanguageactionflowmodel,
      title={$\pi_0$: A Vision-Language-Action Flow Model for General Robot Control}, 
      author={Kevin Black and Noah Brown and Danny Driess and Adnan Esmail and Michael Equi and Chelsea Finn and Niccolo Fusai and Lachy Groom and Karol Hausman and Brian Ichter and Szymon Jakubczak and Tim Jones and Liyiming Ke and Sergey Levine and Adrian Li-Bell and Mohith Mothukuri and Suraj Nair and Karl Pertsch and Lucy Xiaoyang Shi and James Tanner and Quan Vuong and Anna Walling and Haohuan Wang and Ury Zhilinsky},
      year={2026},
      eprint={2410.24164},
      archivePrefix={arXiv},
      primaryClass={cs.LG},
      url={https://arxiv.org/abs/2410.24164}, 
}

@misc{intelligence2025pi05visionlanguageactionmodelopenworld,
      title={$\pi_{0.5}$: a Vision-Language-Action Model with Open-World Generalization}, 
      author={Physical Intelligence and Kevin Black and Noah Brown and James Darpinian and Karan Dhabalia and Danny Driess and Adnan Esmail and Michael Equi and Chelsea Finn and Niccolo Fusai and Manuel Y. Galliker and Dibya Ghosh and Lachy Groom and Karol Hausman and Brian Ichter and Szymon Jakubczak and Tim Jones and Liyiming Ke and Devin LeBlanc and Sergey Levine and Adrian Li-Bell and Mohith Mothukuri and Suraj Nair and Karl Pertsch and Allen Z. Ren and Lucy Xiaoyang Shi and Laura Smith and Jost Tobias Springenberg and Kyle Stachowicz and James Tanner and Quan Vuong and Homer Walke and Anna Walling and Haohuan Wang and Lili Yu and Ury Zhilinsky},
      year={2025},
      eprint={2504.16054},
      archivePrefix={arXiv},
      primaryClass={cs.LG},
      url={https://arxiv.org/abs/2504.16054}, 
}

@misc{kim2025finetuningvisionlanguageactionmodelsoptimizing,
      title={Fine-Tuning Vision-Language-Action Models: Optimizing Speed and Success}, 
      author={Moo Jin Kim and Chelsea Finn and Percy Liang},
      year={2025},
      eprint={2502.19645},
      archivePrefix={arXiv},
      primaryClass={cs.RO},
      url={https://arxiv.org/abs/2502.19645}, 
}

@misc{brohan2023rt1roboticstransformerrealworld,
      title={RT-1: Robotics Transformer for Real-World Control at Scale}, 
      author={Anthony Brohan and Noah Brown and Justice Carbajal and Yevgen Chebotar and Joseph Dabis and Chelsea Finn and Keerthana Gopalakrishnan and Karol Hausman and Alex Herzog and Jasmine Hsu and Julian Ibarz and Brian Ichter and Alex Irpan and Tomas Jackson and Sally Jesmonth and Nikhil J Joshi and Ryan Julian and Dmitry Kalashnikov and Yuheng Kuang and Isabel Leal and Kuang-Huei Lee and Sergey Levine and Yao Lu and Utsav Malla and Deeksha Manjunath and Igor Mordatch and Ofir Nachum and Carolina Parada and Jodilyn Peralta and Emily Perez and Karl Pertsch and Jornell Quiambao and Kanishka Rao and Michael Ryoo and Grecia Salazar and Pannag Sanketi and Kevin Sayed and Jaspiar Singh and Sumedh Sontakke and Austin Stone and Clayton Tan and Huong Tran and Vincent Vanhoucke and Steve Vega and Quan Vuong and Fei Xia and Ted Xiao and Peng Xu and Sichun Xu and Tianhe Yu and Brianna Zitkovich},
      year={2023},
      eprint={2212.06817},
      archivePrefix={arXiv},
      primaryClass={cs.RO},
      url={https://arxiv.org/abs/2212.06817}, 
}

@misc{brohan2023rt2visionlanguageactionmodelstransfer,
      title={RT-2: Vision-Language-Action Models Transfer Web Knowledge to Robotic Control}, 
      author={Anthony Brohan and Noah Brown and Justice Carbajal and Yevgen Chebotar and Xi Chen and Krzysztof Choromanski and Tianli Ding and Danny Driess and Avinava Dubey and Chelsea Finn and Pete Florence and Chuyuan Fu and Montse Gonzalez Arenas and Keerthana Gopalakrishnan and Kehang Han and Karol Hausman and Alexander Herzog and Jasmine Hsu and Brian Ichter and Alex Irpan and Nikhil Joshi and Ryan Julian and Dmitry Kalashnikov and Yuheng Kuang and Isabel Leal and Lisa Lee and Tsang-Wei Edward Lee and Sergey Levine and Yao Lu and Henryk Michalewski and Igor Mordatch and Karl Pertsch and Kanishka Rao and Krista Reymann and Michael Ryoo and Grecia Salazar and Pannag Sanketi and Pierre Sermanet and Jaspiar Singh and Anikait Singh and Radu Soricut and Huong Tran and Vincent Vanhoucke and Quan Vuong and Ayzaan Wahid and Stefan Welker and Paul Wohlhart and Jialin Wu and Fei Xia and Ted Xiao and Peng Xu and Sichun Xu and Tianhe Yu and Brianna Zitkovich},
      year={2023},
      eprint={2307.15818},
      archivePrefix={arXiv},
      primaryClass={cs.RO},
      url={https://arxiv.org/abs/2307.15818}, 
}

@misc{octomodelteam2024octoopensourcegeneralistrobot,
      title={Octo: An Open-Source Generalist Robot Policy}, 
      author={Octo Model Team and Dibya Ghosh and Homer Walke and Karl Pertsch and Kevin Black and Oier Mees and Sudeep Dasari and Joey Hejna and Tobias Kreiman and Charles Xu and Jianlan Luo and You Liang Tan and Lawrence Yunliang Chen and Pannag Sanketi and Quan Vuong and Ted Xiao and Dorsa Sadigh and Chelsea Finn and Sergey Levine},
      year={2024},
      eprint={2405.12213},
      archivePrefix={arXiv},
      primaryClass={cs.RO},
      url={https://arxiv.org/abs/2405.12213}, 
}

@misc{kim2024openvlaopensourcevisionlanguageactionmodel,
      title={OpenVLA: An Open-Source Vision-Language-Action Model}, 
      author={Moo Jin Kim and Karl Pertsch and Siddharth Karamcheti and Ted Xiao and Ashwin Balakrishna and Suraj Nair and Rafael Rafailov and Ethan Foster and Grace Lam and Pannag Sanketi and Quan Vuong and Thomas Kollar and Benjamin Burchfiel and Russ Tedrake and Dorsa Sadigh and Sergey Levine and Percy Liang and Chelsea Finn},
      year={2024},
      eprint={2406.09246},
      archivePrefix={arXiv},
      primaryClass={cs.RO},
      url={https://arxiv.org/abs/2406.09246}, 
}

@misc{zhao2023learningfinegrainedbimanualmanipulation,
      title={Learning Fine-Grained Bimanual Manipulation with Low-Cost Hardware}, 
      author={Tony Z. Zhao and Vikash Kumar and Sergey Levine and Chelsea Finn},
      year={2023},
      eprint={2304.13705},
      archivePrefix={arXiv},
      primaryClass={cs.RO},
      url={https://arxiv.org/abs/2304.13705}, 
}

@misc{embodimentcollaboration2025openxembodimentroboticlearning,
      title={Open X-Embodiment: Robotic Learning Datasets and RT-X Models}, 
      author={Embodiment Collaboration and Abby O'Neill and Abdul Rehman and Abhinav Gupta and Abhiram Maddukuri and Abhishek Gupta and Abhishek Padalkar and Abraham Lee and Acorn Pooley and Agrim Gupta and Ajay Mandlekar and Ajinkya Jain and Albert Tung and Alex Bewley and Alex Herzog and Alex Irpan and Alexander Khazatsky and Anant Rai and Anchit Gupta and Andrew Wang and Andrey Kolobov and Anikait Singh and Animesh Garg and Aniruddha Kembhavi and Annie Xie and Anthony Brohan and Antonin Raffin and Archit Sharma and Arefeh Yavary and Arhan Jain and Ashwin Balakrishna and Ayzaan Wahid and Ben Burgess-Limerick and Beomjoon Kim and Bernhard Schölkopf and Blake Wulfe and Brian Ichter and Cewu Lu and Charles Xu and Charlotte Le and Chelsea Finn and Chen Wang and Chenfeng Xu and Cheng Chi and Chenguang Huang and Christine Chan and Christopher Agia and Chuer Pan and Chuyuan Fu and Coline Devin and Danfei Xu and Daniel Morton and Danny Driess and Daphne Chen and Deepak Pathak and Dhruv Shah and Dieter Büchler and Dinesh Jayaraman and Dmitry Kalashnikov and Dorsa Sadigh and Edward Johns and Ethan Foster and Fangchen Liu and Federico Ceola and Fei Xia and Feiyu Zhao and Felipe Vieira Frujeri and Freek Stulp and Gaoyue Zhou and Gaurav S. Sukhatme and Gautam Salhotra and Ge Yan and Gilbert Feng and Giulio Schiavi and Glen Berseth and Gregory Kahn and Guangwen Yang and Guanzhi Wang and Hao Su and Hao-Shu Fang and Haochen Shi and Henghui Bao and Heni Ben Amor and Henrik I Christensen and Hiroki Furuta and Homanga Bharadhwaj and Homer Walke and Hongjie Fang and Huy Ha and Igor Mordatch and Ilija Radosavovic and Isabel Leal and Jacky Liang and Jad Abou-Chakra and Jaehyung Kim and Jaimyn Drake and Jan Peters and Jan Schneider and Jasmine Hsu and Jay Vakil and Jeannette Bohg and Jeffrey Bingham and Jeffrey Wu and Jensen Gao and Jiaheng Hu and Jiajun Wu and Jialin Wu and Jiankai Sun and Jianlan Luo and Jiayuan Gu and Jie Tan and Jihoon Oh and Jimmy Wu and Jingpei Lu and Jingyun Yang and Jitendra Malik and João Silvério and Joey Hejna and Jonathan Booher and Jonathan Tompson and Jonathan Yang and Jordi Salvador and Joseph J. Lim and Junhyek Han and Kaiyuan Wang and Kanishka Rao and Karl Pertsch and Karol Hausman and Keegan Go and Keerthana Gopalakrishnan and Ken Goldberg and Kendra Byrne and Kenneth Oslund and Kento Kawaharazuka and Kevin Black and Kevin Lin and Kevin Zhang and Kiana Ehsani and Kiran Lekkala and Kirsty Ellis and Krishan Rana and Krishnan Srinivasan and Kuan Fang and Kunal Pratap Singh and Kuo-Hao Zeng and Kyle Hatch and Kyle Hsu and Laurent Itti and Lawrence Yunliang Chen and Lerrel Pinto and Li Fei-Fei and Liam Tan and Linxi "Jim" Fan and Lionel Ott and Lisa Lee and Luca Weihs and Magnum Chen and Marion Lepert and Marius Memmel and Masayoshi Tomizuka and Masha Itkina and Mateo Guaman Castro and Max Spero and Maximilian Du and Michael Ahn and Michael C. Yip and Mingtong Zhang and Mingyu Ding and Minho Heo and Mohan Kumar Srirama and Mohit Sharma and Moo Jin Kim and Muhammad Zubair Irshad and Naoaki Kanazawa and Nicklas Hansen and Nicolas Heess and Nikhil J Joshi and Niko Suenderhauf and Ning Liu and Norman Di Palo and Nur Muhammad Mahi Shafiullah and Oier Mees and Oliver Kroemer and Osbert Bastani and Pannag R Sanketi and Patrick "Tree" Miller and Patrick Yin and Paul Wohlhart and Peng Xu and Peter David Fagan and Peter Mitrano and Pierre Sermanet and Pieter Abbeel and Priya Sundaresan and Qiuyu Chen and Quan Vuong and Rafael Rafailov and Ran Tian and Ria Doshi and Roberto Martín-Martín and Rohan Baijal and Rosario Scalise and Rose Hendrix and Roy Lin and Runjia Qian and Ruohan Zhang and Russell Mendonca and Rutav Shah and Ryan Hoque and Ryan Julian and Samuel Bustamante and Sean Kirmani and Sergey Levine and Shan Lin and Sherry Moore and Shikhar Bahl and Shivin Dass and Shubham Sonawani and Shubham Tulsiani and Shuran Song and Sichun Xu and Siddhant Haldar and Siddharth Karamcheti and Simeon Adebola and Simon Guist and Soroush Nasiriany and Stefan Schaal and Stefan Welker and Stephen Tian and Subramanian Ramamoorthy and Sudeep Dasari and Suneel Belkhale and Sungjae Park and Suraj Nair and Suvir Mirchandani and Takayuki Osa and Tanmay Gupta and Tatsuya Harada and Tatsuya Matsushima and Ted Xiao and Thomas Kollar and Tianhe Yu and Tianli Ding and Todor Davchev and Tony Z. Zhao and Travis Armstrong and Trevor Darrell and Trinity Chung and Vidhi Jain and Vikash Kumar and Vincent Vanhoucke and Vitor Guizilini and Wei Zhan and Wenxuan Zhou and Wolfram Burgard and Xi Chen and Xiangyu Chen and Xiaolong Wang and Xinghao Zhu and Xinyang Geng and Xiyuan Liu and Xu Liangwei and Xuanlin Li and Yansong Pang and Yao Lu and Yecheng Jason Ma and Yejin Kim and Yevgen Chebotar and Yifan Zhou and Yifeng Zhu and Yilin Wu and Ying Xu and Yixuan Wang and Yonatan Bisk and Yongqiang Dou and Yoonyoung Cho and Youngwoon Lee and Yuchen Cui and Yue Cao and Yueh-Hua Wu and Yujin Tang and Yuke Zhu and Yunchu Zhang and Yunfan Jiang and Yunshuang Li and Yunzhu Li and Yusuke Iwasawa and Yutaka Matsuo and Zehan Ma and Zhuo Xu and Zichen Jeff Cui and Zichen Zhang and Zipeng Fu and Zipeng Lin},
      year={2025},
      eprint={2310.08864},
      archivePrefix={arXiv},
      primaryClass={cs.RO},
      url={https://arxiv.org/abs/2310.08864}, 
}

@misc{khazatsky2025droidlargescaleinthewildrobot,
      title={DROID: A Large-Scale In-The-Wild Robot Manipulation Dataset}, 
      author={Alexander Khazatsky and Karl Pertsch and Suraj Nair and Ashwin Balakrishna and Sudeep Dasari and Siddharth Karamcheti and Soroush Nasiriany and Mohan Kumar Srirama and Lawrence Yunliang Chen and Kirsty Ellis and Peter David Fagan and Joey Hejna and Masha Itkina and Marion Lepert and Yecheng Jason Ma and Patrick Tree Miller and Jimmy Wu and Suneel Belkhale and Shivin Dass and Huy Ha and Arhan Jain and Abraham Lee and Youngwoon Lee and Marius Memmel and Sungjae Park and Ilija Radosavovic and Kaiyuan Wang and Albert Zhan and Kevin Black and Cheng Chi and Kyle Beltran Hatch and Shan Lin and Jingpei Lu and Jean Mercat and Abdul Rehman and Pannag R Sanketi and Archit Sharma and Cody Simpson and Quan Vuong and Homer Rich Walke and Blake Wulfe and Ted Xiao and Jonathan Heewon Yang and Arefeh Yavary and Tony Z. Zhao and Christopher Agia and Rohan Baijal and Mateo Guaman Castro and Daphne Chen and Qiuyu Chen and Trinity Chung and Jaimyn Drake and Ethan Paul Foster and Jensen Gao and Vitor Guizilini and David Antonio Herrera and Minho Heo and Kyle Hsu and Jiaheng Hu and Muhammad Zubair Irshad and Donovon Jackson and Charlotte Le and Yunshuang Li and Kevin Lin and Roy Lin and Zehan Ma and Abhiram Maddukuri and Suvir Mirchandani and Daniel Morton and Tony Nguyen and Abigail O'Neill and Rosario Scalise and Derick Seale and Victor Son and Stephen Tian and Emi Tran and Andrew E. Wang and Yilin Wu and Annie Xie and Jingyun Yang and Patrick Yin and Yunchu Zhang and Osbert Bastani and Glen Berseth and Jeannette Bohg and Ken Goldberg and Abhinav Gupta and Abhishek Gupta and Dinesh Jayaraman and Joseph J Lim and Jitendra Malik and Roberto Martín-Martín and Subramanian Ramamoorthy and Dorsa Sadigh and Shuran Song and Jiajun Wu and Michael C. Yip and Yuke Zhu and Thomas Kollar and Sergey Levine and Chelsea Finn},
      year={2025},
      eprint={2403.12945},
      archivePrefix={arXiv},
      primaryClass={cs.RO},
      url={https://arxiv.org/abs/2403.12945}, 
}

@misc{li2026gazevlalearninghumanintention,
      title={GazeVLA: Learning Human Intention for Robotic Manipulation}, 
      author={Chengyang Li and Kaiyi Xiong and Yuan Xu and Lei Qian and Yizhou Wang and Wentao Zhu},
      year={2026},
      eprint={2604.22615},
      archivePrefix={arXiv},
      primaryClass={cs.RO},
      url={https://arxiv.org/abs/2604.22615}, 
}

@misc{pani2026gazeregularizedvisionlanguageactionmodelsrobotic,
      title={Gaze-Regularized Vision-Language-Action Models for Robotic Manipulation}, 
      author={Anupam Pani and Yanchao Yang},
      year={2026},
      eprint={2603.23202},
      archivePrefix={arXiv},
      primaryClass={cs.CV},
      url={https://arxiv.org/abs/2603.23202}, 
}

@misc{pani2026gazeacthumangazeannotatedmanipulationdataset,
      title={GazeAct: A Human-Gaze-Annotated Manipulation Dataset for Attention-Regularized Policies}, 
      author={Anupam Pani and Wang Yuan and Qifan Wang and Yanchao Yang},
      year={2026},
      eprint={2610.24507},
      archivePrefix={arXiv},
      primaryClass={cs.RO},
      url={https://arxiv.org/abs/2610.24507}, 
}

@article{Jiang_2025_kaiwu,
   title={Kaiwu: A Multimodal Manipulation Dataset and Framework for Robot Learning and Human-Robot Interaction},
   volume={10},
   ISSN={2377-3774},
   url={http://dx.doi.org/10.1109/LRA.2025.3609615},
   DOI={10.1109/lra.2025.3609615},
   number={11},
   journal={IEEE Robotics and Automation Letters},
   publisher={Institute of Electrical and Electronics Engineers (IEEE)},
   author={Jiang, Shuo and Li, Haonan and Ren, Ruochen and Zhou, Yanmin and Wang, Zhipeng and He, Bin},
   year={2025},
   month=Nov, pages={11482–11489}
}

@misc{he2026forcebandlearningforcefulmanipulation,
      title={ForceBand: Learning Forceful Manipulation with sEMG}, 
      author={Botao He and Zhi Wang and Linna Kuang and Ishaan Ghosh and Jitendra Malik and Cornelia Fermuller and Tingfan Wu and Jiayuan Mao and Ruoshi Liu and Haozhi Qi and Yiannis Aloimonos},
      year={2026},
      eprint={2606.26093},
      archivePrefix={arXiv},
      primaryClass={cs.RO},
      url={https://arxiv.org/abs/2606.26093}, 
}

@inproceedings{dessalene2026motorsense,
  title     = {{MotorSense}: A Video-{EMG} Dataset of Motoric Representations for Action Understanding},
  author    = {Dessalene, Eadom and Maynord, Michael and Shahidzadeh, Amir-Hossein and He, Botao and Tabatabaee, Yasamin and Liu, Ruoshi and Gao, Ruohan and Liang, Paul Pu and Aloimonos, Yiannis},
  booktitle = {Advances in Neural Information Processing Systems (NeurIPS)},
  year      = {2026},
  url       = {https://arxiv.org/abs/2610.41127}
}

@article{NORMAN1979,
  title = {Electromechanical delay in skeletal muscle under normal movement conditions},
  volume = {106},
  ISSN = {1365-201X},
  url = {http://dx.doi.org/10.1111/j.1748-1716.1979.tb06394.x},
  DOI = {10.1111/j.1748-1716.1979.tb06394.x},
  number = {3},
  journal = {Acta Physiologica Scandinavica},
  publisher = {Wiley},
  author = {NORMAN,  ROBERT W. and KOMI,  PAAVO V.},
  year = {1979},
  month = July,
  pages = {241–248}
}

@inproceedings{Koch2017,
  title = {Early prediction of future hand movements using sEMG data},
  url = {http://dx.doi.org/10.1109/EMBC.2017.8036761},
  DOI = {10.1109/embc.2017.8036761},
  booktitle = {2017 39th Annual International Conference of the IEEE Engineering in Medicine and Biology Society (EMBC)},
  publisher = {IEEE},
  author = {Koch,  Philipp and Phan,  Huy and Maass,  Marco and Katzberg,  Fabrice and Mertins,  Alfred},
  year = {2017},
  month = July,
  pages = {54–57}
}

@article{Koirala2015,
  author  = {Koirala, Kishor and Dasog, Meera
             and Liu, Pu and Clancy, Edward A.},
  title   = {Using the Electromyogram to Anticipate
             Torques About the Elbow},
  journal = {IEEE Transactions on Neural Systems
             and Rehabilitation Engineering},
  volume  = {23},
  number  = {3},
  pages   = {396--402},
  year    = {2015},
  doi     = {10.1109/TNSRE.2014.2331686}
}

@article{Smith2011,
  author  = {Smith, Lauren H. and Hargrove, Levi J.
             and Lock, Blair A. and Kuiken, Todd A.},
  title   = {Determining the Optimal Window Length for
             Pattern Recognition-Based Myoelectric Control:
             Balancing the Competing Effects of
             Classification Error and Controller Delay},
  journal = {IEEE Transactions on Neural Systems
             and Rehabilitation Engineering},
  volume  = {19},
  number  = {2},
  pages   = {186--192},
  year    = {2011},
  doi     = {10.1109/TNSRE.2010.2100828}
}

@misc{zhao2026dexemgdexterousteleoperationemg2pose,
      title={DexEMG: Towards Dexterous Teleoperation System via EMG2Pose Generalization}, 
      author={Qianyou Zhao and Wenqiao Li and Chiyu Wang and Kaifeng Zhang},
      year={2026},
      eprint={2603.05861},
      archivePrefix={arXiv},
      primaryClass={cs.RO},
      url={https://arxiv.org/abs/2603.05861}, 
}

@article{beda2018,
  title = {A Vision-Driven Collaborative Robotic Grasping System Tele-Operated by Surface Electromyography},
  volume = {18},
  ISSN = {1424-8220},
  url = {http://dx.doi.org/10.3390/s18072366},
  DOI = {10.3390/s18072366},
  number = {7},
  journal = {Sensors},
  publisher = {MDPI AG},
  author = {Úbeda,  Andrés and Zapata-Impata,  Brayan S. and Puente,  Santiago T. and Gil,  Pablo and Candelas,  Francisco and Torres,  Fernando},
  year = {2018},
  month = July,
  pages = {2366}
}

@misc{belghazi2021minemutualinformationneural,
      title={MINE: Mutual Information Neural Estimation}, 
      author={Mohamed Ishmael Belghazi and Aristide Baratin and Sai Rajeswar and Sherjil Ozair and Yoshua Bengio and Aaron Courville and R Devon Hjelm},
      year={2021},
      eprint={1801.04062},
      archivePrefix={arXiv},
      primaryClass={cs.LG},
      url={https://arxiv.org/abs/1801.04062}, 
}

@misc{wang2024dexcapscalableportablemocap,
      title={DexCap: Scalable and Portable Mocap Data Collection System for Dexterous Manipulation}, 
      author={Chen Wang and Haochen Shi and Weizhuo Wang and Ruohan Zhang and Li Fei-Fei and C. Karen Liu},
      year={2024},
      eprint={2403.07788},
      archivePrefix={arXiv},
      primaryClass={cs.RO},
      url={https://arxiv.org/abs/2403.07788}, 
}

@article{tao2025dexwild,
      title={DexWild: Dexterous Human Interactions for In-the-Wild Robot Policies},
      author={Tao, Tony and Srirama, Mohan Kumar and Liu, Jason Jingzhou and Shaw, Kenneth and Pathak, Deepak},
      journal={Robotics: Science and Systems (RSS)},
      year={2025}}

@misc{yu2025mimictouchleveragingmultimodalhuman,
      title={MimicTouch: Leveraging Multi-modal Human Tactile Demonstrations for Contact-rich Manipulation}, 
      author={Kelin Yu and Yunhai Han and Qixian Wang and Vaibhav Saxena and Danfei Xu and Ye Zhao},
      year={2025},
      eprint={2310.16917},
      archivePrefix={arXiv},
      primaryClass={cs.RO},
      url={https://arxiv.org/abs/2310.16917}, 
}

@ARTICLE{204774,
  author={Hudgins, B. and Parker, P. and Scott, R.N.},
  journal={IEEE Transactions on Biomedical Engineering}, 
  title={A new strategy for multifunction myoelectric control}, 
  year={1993},
  volume={40},
  number={1},
  pages={82-94},
  doi={10.1109/10.204774}}

@ARTICLE{6737308,
  author={Farina, Dario and Jiang, Ning and Rehbaum, Hubertus and Holobar, Aleš and Graimann, Bernhard and Dietl, Hans and Aszmann, Oskar C.},
  journal={IEEE Transactions on Neural Systems and Rehabilitation Engineering}, 
  title={The Extraction of Neural Information from the Surface EMG for the Control of Upper-Limb Prostheses: Emerging Avenues and Challenges}, 
  year={2014},
  volume={22},
  number={4},
  pages={797-809},
  doi={10.1109/TNSRE.2014.2305111}}

@ARTICLE{10.3389/fnins.2021.694914,
AUTHOR={Guan, Yuan  and Wang, Ning  and Yang, Chenguang },         
TITLE={An Improvement of Robot Stiffness-Adaptive Skill Primitive Generalization Using the Surface Electromyography in Human–Robot Collaboration},
          
JOURNAL={Frontiers in Neuroscience},
          
VOLUME={Volume 15 - 2021},
  
YEAR={2021},
  
URL={https://www.frontiersin.org/journals/neuroscience/articles/10.3389/fnins.2021.694914},
  
DOI={10.3389/fnins.2021.694914},
  
ISSN={1662-453X}}

@article{ZHANG2022330,
title = {An electromyography signals-based human-robot collaboration method for human skill learning and imitation},
journal = {Journal of Manufacturing Systems},
volume = {64},
pages = {330-343},
year = {2022},
issn = {0278-6125},
doi = {https://doi.org/10.1016/j.jmsy.2022.07.005},
url = {https://www.sciencedirect.com/science/article/pii/S0278612522001170},
author = {Tie Zhang and Hanlei Sun and Yanbiao Zou and Hubo Chu}
}

@ARTICLE{10.3389/fnins.2021.733359,
    
AUTHOR={Kim, Keun-Tae  and Park, Sangsoo  and Lim, Tae-Hyun  and Lee, Song Joo },
           
TITLE={Upper-Limb Electromyogram Classification of Reaching-to-Grasping Tasks Based on Convolutional Neural Networks for Control of a Prosthetic Hand},
          
JOURNAL={Frontiers in Neuroscience},
          
VOLUME={Volume 15 - 2021},
  
YEAR={2021},
  
URL={https://www.frontiersin.org/journals/neuroscience/articles/10.3389/fnins.2021.733359},
  
DOI={10.3389/fnins.2021.733359},
  
ISSN={1662-453X}}

@article{HERMENS2000361,
title = {Development of recommendations for SEMG sensors and sensor placement procedures},
journal = {Journal of Electromyography and Kinesiology},
volume = {10},
number = {5},
pages = {361-374},
year = {2000},
issn = {1050-6411},
doi = {https://doi.org/10.1016/S1050-6411(00)00027-4},
url = {https://www.sciencedirect.com/science/article/pii/S1050641100000274},
author = {Hermie J Hermens and Bart Freriks and Catherine Disselhorst-Klug and Günter Rau}
}

@inproceedings{NEURIPS2022_b89d5e20,
 author = {Huang, Po-Yao and Xu, Hu and Li, Juncheng and Baevski, Alexei and Auli, Michael and Galuba, Wojciech and Metze, Florian and Feichtenhofer, Christoph},
 booktitle = {Advances in Neural Information Processing Systems},
 doi = {10.52202/068431-2081},
 editor = {S. Koyejo and S. Mohamed and A. Agarwal and D. Belgrave and K. Cho and A. Oh},
 pages = {28708--28720},
 publisher = {Curran Associates, Inc.},
 title = {Masked Autoencoders that Listen},
 url = {https://proceedings.neurips.cc/paper_files/paper/2022/file/b89d5e209990b19e33b418e14f323998-Paper-Conference.pdf},
 volume = {35},
 year = {2022}
}

@ARTICLE{Yang2025-eh,
  title     = "Calibration-free {sEMG} intention recognition via
               self-supervised pretraining and adversarial domain alignment for
               upper-limb rehabilitation",
  author    = "Yang, Yuanbo and Teo, Hiu Hong and King, Yeong Jin and Zhang,
               Yao and Wang, Gang and Qu, Xiangxu",
  journal   = "Sci. Rep.",
  publisher = "Springer Science and Business Media LLC",
  volume    =  15,
  number    =  1,
  pages     = "44850",
  month     =  dec,
  year      =  2025,
  copyright = "https://creativecommons.org/licenses/by-nc-nd/4.0",
  language  = "en"
}

@article{RAGHU2025109479,
title = {Self-supervised learning via VICReg enables training of EMG pattern recognition using continuous data with unclear labels},
journal = {Computers in Biology and Medicine},
volume = {185},
pages = {109479},
year = {2025},
issn = {0010-4825},
doi = {https://doi.org/10.1016/j.compbiomed.2024.109479},
url = {https://www.sciencedirect.com/science/article/pii/S0010482524015646},
author = {Shriram Tallam Puranam Raghu and Dawn T. MacIsaac and Erik J. Scheme}
}

@ARTICLE{biagiotti,
  author={Biagiotti, Luigi and Meattini, Roberto and Chiaravalli, Davide and Palli, Gianluca and Melchiorri, Claudio},
  journal={IEEE Transactions on Robotics}, 
  title={Robot Programming by Demonstration: Trajectory Learning Enhanced by sEMG-Based User Hand Stiffness Estimation}, 
  year={2023},
  volume={39},
  number={4},
  pages={3259-3278},
  doi={10.1109/TRO.2023.3258669}}

@article{LIU2024127248,
title = {Robot skill learning system of multi-space fusion based on dynamic movement primitives and adaptive neural network control},
journal = {Neurocomputing},
volume = {574},
pages = {127248},
year = {2024},
issn = {0925-2312},
doi = {https://doi.org/10.1016/j.neucom.2024.127248},
url = {https://www.sciencedirect.com/science/article/pii/S0925231224000195},
author = {Chengguo Liu and Guangzhu Peng and Yu Xia and Junyang Li and Chenguang Yang}
}

@Article{biomimetics10060399,
AUTHOR = {Xia, Wei and Liao, Zhiwei and Lu, Zongxin and Yao, Ligang},
TITLE = {Bio-Signal-Guided Robot Adaptive Stiffness Learning via Human-Teleoperated Demonstrations},
JOURNAL = {Biomimetics},
VOLUME = {10},
YEAR = {2025},
NUMBER = {6},
ARTICLE-NUMBER = {399},
URL = {https://www.mdpi.com/2313-7673/10/6/399},
PubMedID = {40558368},
ISSN = {2313-7673},
DOI = {10.3390/biomimetics10060399}
}

@article{Wang2025HumanRobotIS,
  title={Human–Robot Intrinsic Skill Transfer and Programming by Demonstration System},
  author={Fei Wang and Jinxiu Wu and Siyi Lian and Yi Guo and Kaiyin Hu},
  journal={IEEE Transactions on Automation Science and Engineering},
  year={2025},
  volume={22},
  pages={14498-14509},
  url={https://api.semanticscholar.org/CorpusID:277716346}
}

@article{Kraskov_2004,
   title={Estimating mutual information},
   volume={69},
   ISSN={1550-2376},
   url={http://dx.doi.org/10.1103/PhysRevE.69.066138},
   DOI={10.1103/physreve.69.066138},
   number={6},
   journal={Physical Review E},
   publisher={American Physical Society (APS)},
   author={Kraskov, Alexander and Stögbauer, Harald and Grassberger, Peter},
   year={2004},
   month=June }

@misc{MINE,
      title={MINE: Mutual Information Neural Estimation}, 
      author={Mohamed Ishmael Belghazi and Aristide Baratin and Sai Rajeswar and Sherjil Ozair and Yoshua Bengio and Aaron Courville and R Devon Hjelm},
      year={2021},
      eprint={1801.04062},
      archivePrefix={arXiv},
      primaryClass={cs.LG},
      url={https://arxiv.org/abs/1801.04062}, 
}

@article{GCMI1,
author = {Ince, Robin A.A. and Giordano, Bruno L. and Kayser, Christoph and Rousselet, Guillaume A. and Gross, Joachim and Schyns, Philippe G.},
title = {A statistical framework for neuroimaging data analysis based on mutual information estimated via a gaussian copula},
journal = {Human Brain Mapping},
volume = {38},
number = {3},
pages = {1541-1573},
doi = {https://doi.org/10.1002/hbm.23471},
url = {https://onlinelibrary.wiley.com/doi/abs/10.1002/hbm.23471},
eprint = {https://onlinelibrary.wiley.com/doi/pdf/10.1002/hbm.23471},
year = {2017}
}

@misc{SMILE,
      title={Understanding the Limitations of Variational Mutual Information Estimators}, 
      author={Jiaming Song and Stefano Ermon},
      year={2020},
      eprint={1910.06222},
      archivePrefix={arXiv},
      primaryClass={cs.LG},
      url={https://arxiv.org/abs/1910.06222}, 
}
}

\clearpage
\appendix

\textbf{Appendix Overview.}
This Appendix provides additional details on
Robo-EMG data collection, EMG-action temporal
analysis, supervision target construction,
and VLA training. We further describe the
experimental protocols, visual attention
analysis, and additional EMG-action association
controls. For ease of reference,
Table~\ref{tab:app_notation} summarizes
the notation used throughout the paper.

\begin{table}[h]
    \centering
    \caption{
        Summary of the main notation used
        in Robo-EMG.
    }
    \label{tab:app_notation}
    \footnotesize
    \setlength{\tabcolsep}{3pt}
    \renewcommand{\arraystretch}{1.15}

    \begin{tabularx}{\columnwidth}{@{}lX@{}}
        \toprule
        \textbf{Symbol} & \textbf{Description} \\
        \midrule

        $\mathbf{a}_t \in \mathbb{R}^{7}$
        & Single-arm robot action at time $t$ \\

        $\mathbf{A}_t \in \mathbb{R}^{H\times7}$
        & Action chunk of horizon $H$ \\

        $X_{W,\Delta}(t)$
        & Eight-channel EMG window associated
          with action at $t$ \\

        $W,\Delta$
        & EMG window duration and temporal offset \\

        $N_W$
        & Number of EMG samples in a window \\

        $(W^\star,\Delta^\star)$
        & Selected EMG--action temporal alignment \\

        $\mathbf{e}_t^{W,\Delta} \in \mathbb{R}^{8}$
        & Sensor-wise EMG activation vector \\

        $\mathbf{y}_t^{\mathrm{act}} \in \mathbb{R}^{8}$
        & EMG Activation supervision target \\

        $\mathbf{z}_t^{\mathrm{EMG}} \in \mathbb{R}^{128}$
        & Learned EMG embedding target \\

        $\mathbf{h}_t$
        & Pooled final-layer VLM features \\

        $\lambda_{\mathrm{EMG}}$
        & Auxiliary EMG loss weight \\

        \bottomrule
    \end{tabularx}
\end{table}

\section{Data Collection and Synchronization Details}
\label{app:data_collection}

\subsection{Participants and Demonstration Allocation}
\label{app:participants}

Robo-EMG comprises of 5,500 human-teleoperated
demonstrations collected across real-world
and simulated manipulation environments.
In the real-world setting, ten teleoperators
used the ALOHA leader-follower platform
to perform 15 single-arm manipulation tasks.
Each teleoperator contributed 30 demonstrations
per task, yielding 300 demonstrations per task
and 4,500 demonstrations in total.

For simulated data collection, ten human
teleoperators controlled a simulated robot
in MuJoCo using the physical ALOHA leader
interface. We collected 50 demonstrations
per task across 20 manipulation tasks,
resulting in 1,000 original demonstrations.
Through visual augmentation, these are
expanded to 30,000 trajectory instances,
without requiring additional human
demonstrations. The augmentation procedure
is detailed in Sec.~\ref{app:data_augmentation}.

In both settings, robot actions originate
from human teleoperation, and EMG signals
are recorded directly from the operators
rather than synthetically generated.

\subsection{EMG Acquisition and Cross-Modal Synchronization}
\label{app:data_sync}

Surface EMG was recorded using a Delsys
Trigno system with eight channels sampled
at 500~Hz. Sensors were placed across the
hand, forearm, and upper arm to capture
muscle activity associated with grasping,
wrist movement, and arm manipulation.
The recording sites and teleoperation
setup are illustrated in
Fig.~\ref{fig:equipment-used}.

For real-world demonstrations, hardware
transistor-transistor logic (TTL) markers
were recorded by both the robot and EMG
acquisition systems at the beginning and
end of each trial. These shared events,
together with acquisition timestamps,
were used to align the recorded streams
on a common timeline during offline
processing. This alignment allows robot
actions at time $t$ to be associated with
corresponding EMG segments, providing
the basis for the temporal window and
offset analysis described in
Appendix~\ref{app:emg_temporal}.

Importantly, synchronization establishes
a common temporal reference but does not
assume that muscle activity and robot
actions occur simultaneously. We therefore also
analyze different EMG window durations
and temporal offsets to identify
informative EMG-action relationships.

\subsection{Simulation Data Augmentation}
\label{app:data_augmentation}

To improve visual diversity, we augment
the 1,000 human-teleoperated simulation
demonstrations by replaying their recorded
trajectories under varied scene appearances,
expanding the dataset to approximately
30,000 trajectory instances. The augmentation
introduces changes to backgrounds and surface
textures, repositions task-irrelevant objects,
and adds environmental clutter. We further
randomize object materials, lighting intensity
and color, and camera viewpoints to introduce
additional visual variation.

Throughout augmentation, the original
robot actions, task-relevant object motions,
and recorded human EMG signals are preserved,
maintaining their temporal correspondence.
Scene variations are constrained to avoid
interference with task execution and preserve
the visibility of task-relevant objects.
During policy training, additional image-level
augmentations, including random cropping,
rotation, and color jitter, are applied.

\begin{figure}[t]
    \centering
    \includegraphics[width=\columnwidth]{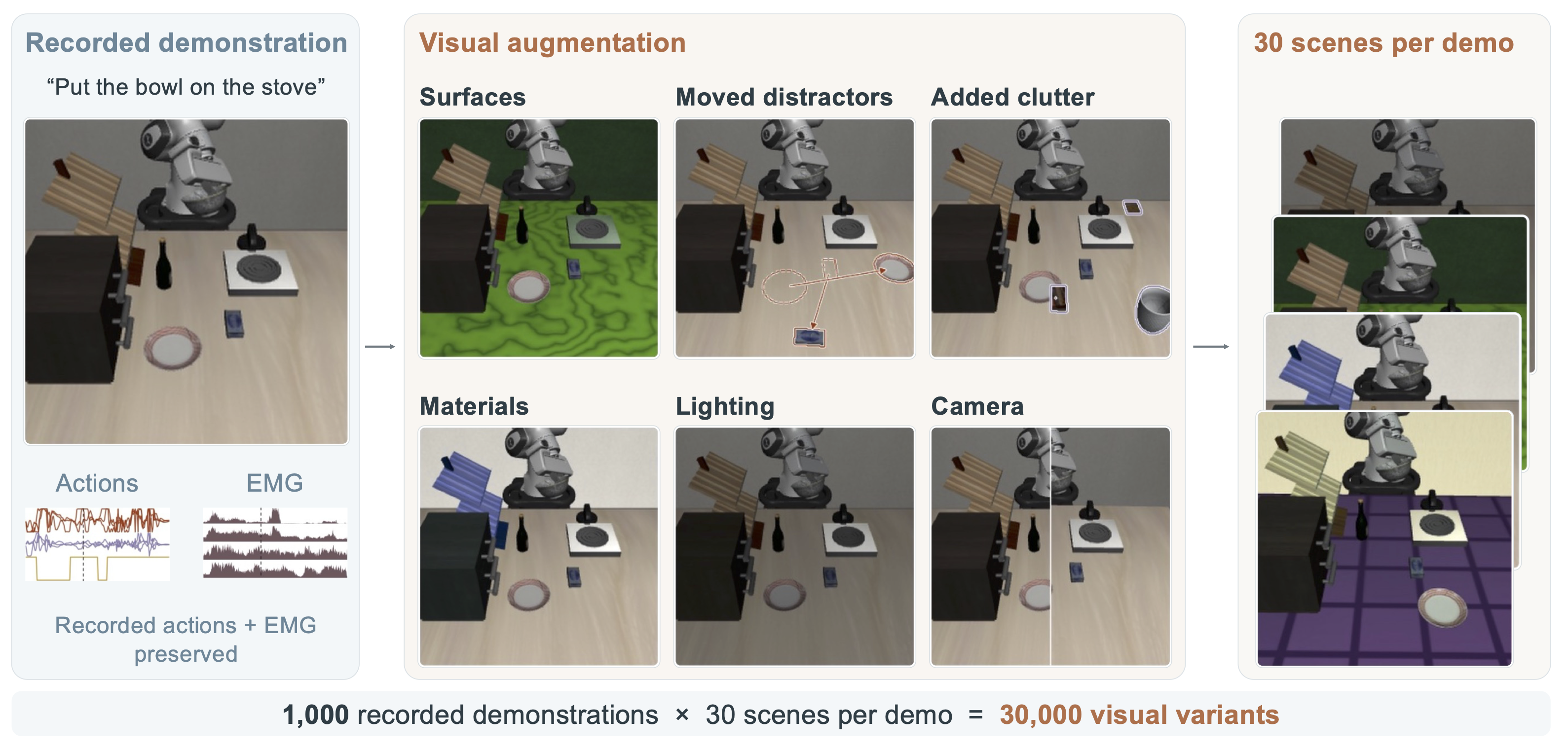}
    \caption{
        \textbf{Visual augmentation of simulated demonstrations.}
        Each of the 1,000 human-teleoperated
        demonstrations is replayed across
        30 visually varied scenes, yielding
        30,000 trajectory instances.
        Scene appearance, lighting, textures,
        and background elements are varied
        while preserving the recorded robot
        actions and corresponding human EMG.
    }
    \label{fig:robo_emg_augmentation_simulation}
\end{figure}

Importantly, these augmented trajectories
represent visual variations of the original
human demonstrations rather than newly
collected teleoperation episodes.
No synthetic EMG signals are introduced.
Figure~\ref{fig:robo_emg_augmentation_simulation}
illustrates the augmentation process
and representative scene variations.

\subsection{Participant Preparation and EMG Quality Control}
\label{app:ethics}

The study received prior institutional ethics
approval, and informed consent was obtained
from all participating teleoperators.
Before data collection, participants were
briefed on the experimental objectives,
teleoperation setup, and task procedures.
They were first allowed to practice using
the ALOHA leader interface until they
had sufficient familiarity and
control to perform the assigned tasks.

After getting familiarized with the procedure,
participants were introduced to the EMG
recording system, including its purpose
and how muscle activity is measured.
The skin at each recording site was
cleaned using rubbing alcohol before
attaching the EMG sensors. Participants
then performed basic movements, including
grasping, releasing, and forward-backward
arm motions, to verify that the recorded
signals exhibited appropriate muscle
activation patterns without excessive noise.

Throughout the experiments, EMG channels
were monitored to identify potential
signal degradation, sensor displacement,
or excessive recording noise. After sensor
placement and signal verification,
participants completed additional practice
attempts to ensure that the sensors did
not interfere with task execution.
Formal demonstration collection began
only after satisfactory teleoperation
performance and EMG signal quality
were confirmed.

\section{EMG-Action Temporal Dependence}
\label{app:emg_temporal}

This section provides implementation details
for the EMG-action temporal analysis introduced
in Sec.~\ref{sec:temporal_dependence}.
We describe signal preprocessing, mutual-information
estimation, consensus-based alignment selection,
and held-out action-prediction probes.

\subsection{Signal Preprocessing and Sensor Activation}
\label{app:emg_preprocessing}

Raw EMG recordings are preprocessed offline
to reduce noise and obtain consistent
muscle-activation estimates. Each of the
eight channels is first median-centered,
followed by a 20-200~Hz band-pass filter
and a 60~Hz notch filter to suppress
out-of-band noise and power-line interference.
The filtered signals are then rectified
and smoothed using a 100~ms moving-average
window to obtain EMG envelopes.

For each robot action $\mathbf{a}_t$
at time $t$, we extract the corresponding
EMG window $X_{W,\Delta}(t)$ as defined
in Eq.~\ref{eq:emg_window}. We summarize
this window using the activation function
$f_{\mathrm{act}}$:

\begin{equation}
    \mathbf{e}_t^{W,\Delta}
    =
    f_{\mathrm{act}}
    \left(X_{W,\Delta}(t)\right)
    \in \mathbb{R}^{8},
    \label{eq:app_sensor_activation}
\end{equation}

where $f_{\mathrm{act}}$ computes the
normalized mean envelope of each EMG
channel. Specifically, for channel $c$,

\begin{equation}
    e_{t,c}^{W,\Delta}
    =
    \frac{
        \overline{g}_c^{W,\Delta}(t)
    }{
        Q_{0.95}(g_c;\mathcal{T}_r)
    },
    \label{eq:app_activation_component}
\end{equation}

where $\overline{g}_c^{W,\Delta}(t)$
denotes the mean processed EMG envelope
of channel $c$ within the window
$[t-\Delta-W,\;t-\Delta]$.
The denominator
$Q_{0.95}(g_c;\mathcal{T}_r)$
is the empirical 95th percentile of
channel $c$'s envelope, computed over
the entire recording $r$.
This recording-level reference is shared
across all windows from the same trial.

Applying this operation independently
to all eight channels produces a
sensor-wise activation vector that
preserves relative muscle activity
across channels. 

Windows containing fewer than two EMG
samples or less than 80\% temporal
coverage are excluded from the analysis.
All preprocessing and normalization
are performed offline before temporal
dependence estimation.

\subsection{Mutual-Information Estimation}
\label{app:mi_estimation}

We estimate the statistical dependence
between EMG activation and robot actions
across different temporal configurations.
For each candidate $(W,\Delta)$, the
eight-dimensional activation vector
$\mathbf{e}_t^{W,\Delta}$ is paired with
the corresponding robot action
$\mathbf{a}_t\in\mathbb{R}^{7}$ using
synchronized timestamps. To ensure
consistent comparisons, the same eligible
action timestamps are used across all
evaluated temporal configurations.

We employ four complementary mutual-information
estimators, each capturing statistical
dependence through a different approach:

\begin{itemize}
    \item \textbf{KSG}~\cite{Kraskov_2004}:
    A non-parametric estimator based on
    nearest-neighbor statistics in the
    joint and marginal feature spaces.

    \item \textbf{MINE}~\cite{belghazi2021minemutualinformationneural}:
    A neural estimator that learns a
    variational lower bound on mutual
    information using a trainable critic.

    \item \textbf{GCMI}~\cite{GCMI1}:
    A Gaussian-based estimator
    that uses rank transformations to
    estimate statistical dependence.

    \item \textbf{SMILE}~\cite{SMILE}:
    A variational estimator employing
    clipped density-ratio estimates to
    reduce estimation variance.
\end{itemize}

Because these estimators rely on different
assumptions and may produce estimates
on different numerical scales, we avoid
directly averaging their MI values.
Instead, for each estimator, candidate
$(W,\Delta)$ configurations are ranked
according to their estimated MI and
converted into percentile ranks
$r_k(W,\Delta)\in[0,1]$.
These normalized ranks are combined
using the consensus score defined in
Eq.~\ref{eq:mi_consensus}.

This rank-based aggregation reduces
dependence on the absolute scale of
any individual estimator and prioritizes
temporal configurations exhibiting
consistently strong EMG--action dependence
across estimation methods.

\subsection{Consensus-Based Temporal Selection}
\label{app:mi_consensus}

To identify informative EMG-action
alignments, we aggregate the percentile
ranks from the four MI estimators using
the consensus score $S(W,\Delta)$ defined
in Eq.~\ref{eq:mi_consensus}.
Configurations receiving consistently
high rankings across estimators achieve
higher consensus scores, reducing reliance
on any single estimation method.

Rather than selecting only the configuration
with the maximum consensus score, we
identify a neighborhood of similarly
informative temporal alignments.
Let $S_{\max}$ denote the maximum
consensus score over the evaluated
$(W,\Delta)$ grid. We define the
high-consensus candidate set as
\begin{equation}
    \mathcal{C}_{\eta}
    =
    \left\{
    (W,\Delta):
    S(W,\Delta)\geq
    (1-\eta)S_{\max}
    \right\},
    \label{eq:app_consensus_region}
\end{equation}
where $\eta$ controls the relative
tolerance from the maximum score. In our work, we chose $\eta=0.1$

We retain the connected region of
$\mathcal{C}_{\eta}$ containing the
consensus maximum, considering neighboring
configurations on the evaluated temporal
grid. This favors a locally consistent
region of strong EMG--action dependence
rather than an isolated high-scoring
configuration that may be sensitive
to estimation variability.

The resulting region provides candidate
alignments for further evaluation using
held-out action-prediction probes.
These probes serve as supporting evidence
when selecting the final temporal
configuration, as described in the
following subsections.

\subsection{Predictive Probes and Final Alignment Selection}
\label{app:predictive_probes}
\label{app:alignment_selection}

To complement the MI-based temporal analysis, we evaluate
whether candidate EMG windows support robot-action prediction
on held-out demonstrations. We use two complementary
predictive probes: a feature-based multilayer perceptron
(MLP) and a waveform-based one-dimensional convolutional
neural network (1D CNN).

For the MLP, we extract four statistics per EMG channel:
root-mean-square amplitude, mean absolute value, standard
deviation, and mean absolute difference between consecutive
samples. The resulting 32-dimensional feature vector is
standardized and passed directly to the MLP.
The 1D CNN instead processes resampled eight-channel EMG
waveforms, learning temporal features directly from the
signals.

For each window duration and offset, both probes predict
the corresponding robot action $a_t$. We use episode-level
cross-validation, with input and target standardization
fitted only on the training episodes of each fold.
Performance on held-out episodes is measured using
variance-weighted coefficient of determination ($R^2$)
and normalized mean-squared error (NMSE), averaged across
folds. These predictive results provide a complementary
check on the temporal structure identified by MI estimation.

\textbf{Final alignment selection.}
MI consensus remains the primary criterion
for identifying informative temporal
regions. The probes provide additional
evidence when choosing among configurations
within the high-consensus region identified
in Sec.~\ref{app:mi_consensus}. 

Following this procedure, we select a
task-specific alignment $(W^\star,\Delta^\star)$
and use the corresponding EMG window
$X_{W^\star,\Delta^\star}(t)$ to construct
the auxiliary supervision targets described
in Sec.~\ref{app:emg_targets}.
The predictive probes are used only for
alignment selection, not VLA training.
Moreover, the selected offsets represent
empirical EMG-action relationships.

\begin{figure*}[t]
    \centering
    \includegraphics[
        width=\textwidth,
        keepaspectratio
    ]{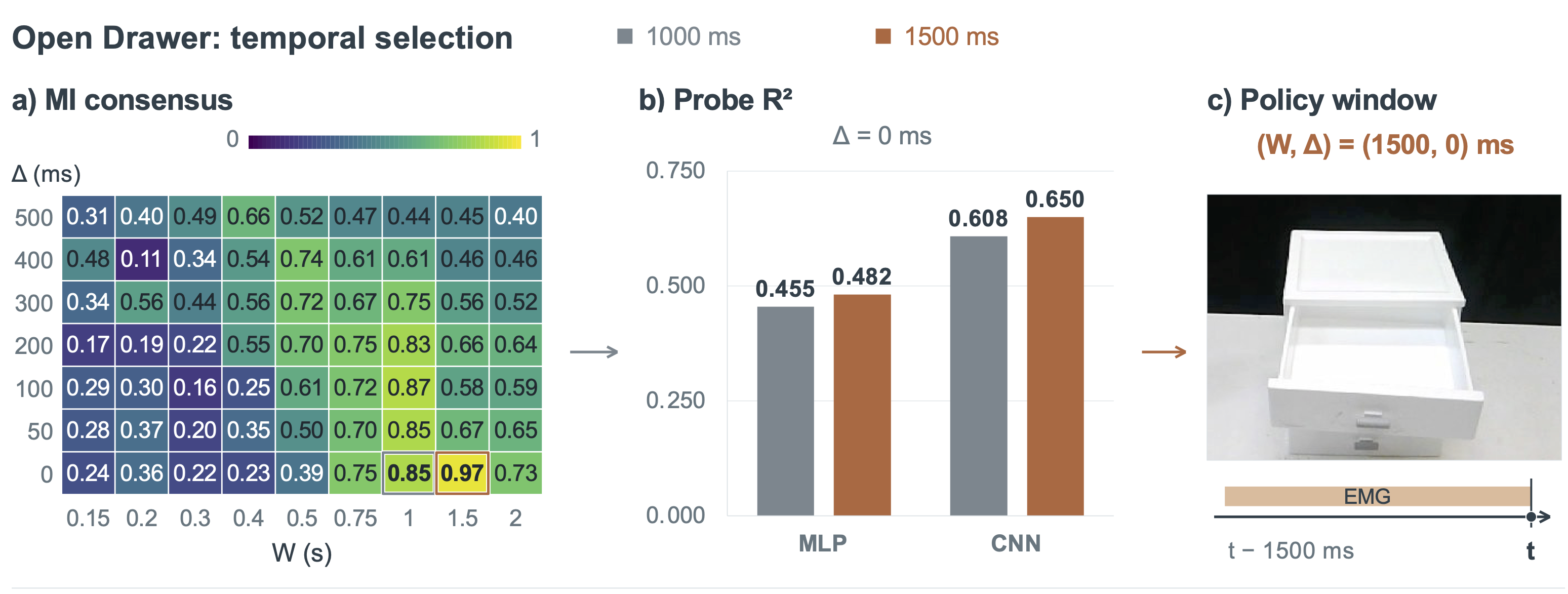}
    \caption{
        \textbf{MI-guided temporal alignment for the
        Open Top Drawer task.}
        (a) Consensus scores across EMG window
        durations $W$ and temporal offsets
        $\Delta$, with the highest score
        ($0.97$) occurring at $(1500,0)$~ms.
        (b) Held-out action-prediction probes
        at $\Delta=0$~ms further support
        this selection: increasing the window
        from 1000 to 1500~ms improves
        $R^2$ from 0.455 to 0.482 for the
        MLP and from 0.608 to 0.650 for
        the 1D CNN.
        (c) The selected configuration
        associates each robot action with
        the preceding 1500~ms of EMG.
        Together, these analyses illustrate
        how MI consensus and predictive
        evidence guide the construction
        of EMG supervision targets.
    }
    \label{fig:app_temporal_dependence}
\end{figure*}

\textbf{Example of temporal alignment.}
Figure~\ref{fig:app_temporal_dependence}
illustrates this selection procedure
using the Open Top Drawer task.
The MI-consensus analysis identifies
$(W,\Delta)=(1500,0)$~ms as the
highest-scoring configuration, with
a consensus score of 0.97.
Importantly, the predictive probes
provide additional support for this
choice. At zero offset, increasing
the window duration from 1000 to
1500~ms improves held-out action
prediction for both the MLP and
1D CNN, suggesting that the longer
EMG history contains additional
action-relevant information.

The selected configuration therefore
uses the preceding 1.5 seconds of
muscle activity to construct the
EMG supervision target for each
robot action. This basically connects
our temporal-dependence analysis from the previous paragraph 
to the task-specific alignment
used in the downstream manipulation
experiments.

\section{EMG Supervision Target Construction}
\label{app:emg_targets}

We construct two alternative EMG supervision
targets from the temporally aligned signals:
sensor-wise EMG activation and learned EMG
embeddings. Both representations are derived
from the selected EMG windows but capture
different levels of information about
human muscle activity.

\subsection{Sensor-Wise EMG Activation}
\label{app:activation_target}

Our first supervision target directly
represents the activation levels recorded
by the eight EMG sensors. For each robot
action at time $t$, we extract the
corresponding EMG window using the selected
alignment $(W^\star,\Delta^\star)$ and
compute the activation vector as

\begin{equation}
    \mathbf{y}_t^{\mathrm{act}}
    =
    f_{\mathrm{act}}
    \left(X_{W^\star,\Delta^\star}(t)\right)
    =
    \mathbf{e}_t^{W^\star,\Delta^\star}
    \in \mathbb{R}^{8}.
    \label{eq:app_activation_target}
\end{equation}

As detailed in Sec.~\ref{app:emg_preprocessing},
each component corresponds to the normalized
mean EMG envelope of an individual sensor.
Normalization is performed independently
using the recording-level 95th-percentile
reference, preserving the relative activation
magnitude of each channel.

This representation provides a simple,
interpretable supervision target that
retains sensor-specific muscle activity.
However, summarizing each EMG window
through mean activation discards finer
temporal and spectral patterns, motivating
the learned EMG representation described
in the following subsection.

\subsection{Masked-Spectrogram EMG Encoder}
\label{app:emg_encoder}

While sensor-wise activation captures the
average activity of individual muscles,
it discards temporal and frequency-dependent
patterns within the EMG signal. To capture
richer physiological information, we
pretrain a Transformer-based masked
spectrogram autoencoder using recorded
EMG signals without robot-action labels.

\textbf{Spectrogram representation.}
Given an eight-channel EMG window
$X_{W,\Delta}(t)$, we apply the short-time
Fourier transform (STFT) independently
to each channel. The resulting complex
spectrograms are converted into
log-power representations:

\begin{equation}
    \mathbf{S}_t^{(c)}
    =
    \log\left(
    \left|\operatorname{STFT}
    (X_{W,\Delta}^{(c)}(t))\right|^2
    + \epsilon
    \right),
    \label{eq:app_emg_spectrogram}
\end{equation}

where $c\in\{1,\ldots,8\}$ indexes the
EMG channels and $\epsilon>0$ ensures
numerical stability. The resulting
spectrogram preserves both temporal
and frequency-dependent variations
in muscle activity.

\textbf{Self-supervised pretraining.}
We train the EMG encoder using masked
spectrogram reconstruction. During
pretraining, portions of the input
spectrogram are masked, and a Transformer
encoder learns representations from the
remaining visible information. A
reconstruction decoder then predicts
the masked spectrogram content.
This objective encourages the encoder
to capture dependencies across temporal
and spectral regions, rather than relying
solely on average signal amplitude.

Importantly, encoder pretraining does
not require robot actions or task labels,
allowing the representation to be learned
directly from the recorded EMG signals.

\textbf{Embedding extraction.}
After pretraining, we retain the encoder
and project its learned features into
a 128-dimensional embedding. For the
selected temporal alignment
$(W^\star,\Delta^\star)$, the supervision
target is obtained as

\begin{equation}
    \mathbf{z}_t^{\mathrm{EMG}}
    =
    f_{\phi}\left(
    X_{W^\star,\Delta^\star}(t)
    \right)
    \in \mathbb{R}^{128},
    \label{eq:app_emg_encoder_target}
\end{equation}

where $f_{\phi}$ includes spectrogram
preprocessing, the pretrained encoder,
and the embedding projection.

The pretrained EMG encoder is frozen
during VLA training. Its embeddings
serve as prediction targets for the
auxiliary prediction head attached
to the VLM's multimodal features.
Consequently, the VLA learns from
the physiological representations
during training without requiring
EMG measurements at inference.

\section{VLA Auxiliary Supervision and Training Details}
\label{app:vla_training}

This section provides implementation details
for integrating EMG supervision into VLA
training, including feature extraction,
auxiliary prediction, and joint optimization
with the action-learning objective.

\subsection{VLM Feature Pooling and Auxiliary Prediction Head}
\label{app:aux_head}

The auxiliary EMG prediction head operates
on the final-layer multimodal prefix features
of the VLM. These representations contain
image tokens from the available camera
observations and tokens from the  language instruction.
Rather than supervising individual tokens,
we aggregate the valid prefix features
into a single representation per observation.

Let $\mathbf{H}_t^{(L)}
\in\mathbb{R}^{N\times d_h}$
denote the final-layer prefix features,
where $N$ is the number of tokens and
$d_h$ is the hidden dimension.
We apply masked mean pooling:

\begin{equation}
    \mathbf{h}_t =
    \frac{
        \sum_{i=1}^{N}
        m_{t,i}\mathbf{H}_{t,i}^{(L)}
    }{
        \max\left(\sum_{i=1}^{N}m_{t,i},1\right)
    },
    \label{eq:app_masked_pooling}
\end{equation}

where $m_{t,i}\in\{0,1\}$ indicates
whether token $i$ is valid, excluding
masked image tokens and prompt padding.
The resulting feature
$\mathbf{h}_t\in\mathbb{R}^{d_h}$
summarizes the available visual and
language information.

This pooled representation is passed
through a  multilayer perceptron
(MLP) with a 256-dimensional hidden layer. The output dimension
depends on the EMG supervision target:
8 for sensor-wise activation, and
128 for the learned EMG embedding.

Importantly, the auxiliary head is attached
to the VLM's prefix representations rather
than the action expert's suffix features,
which are used separately for action
prediction. This design allows EMG
supervision to directly influence the
multimodal features that condition
the action expert, without modifying
its action-prediction architecture.

\subsection{Joint Objective and Training Procedure}
\label{app:joint_objective}

Our training framework combines two
complementary learning objectives: the
standard action prediction objective teaches
the robot how to perform a task, while
auxiliary EMG supervision encourages its
visual-language representations to capture
information associated with the human
demonstrator's muscle activity. We jointly
optimize these objectives as

\begin{equation}
    \mathcal{L}_{\mathrm{total}}
    =
    \mathcal{L}_{\mathrm{action}}
    +
    \lambda_{\mathrm{EMG}}
    \mathcal{L}_{\mathrm{EMG}},
    \label{eq:app_joint_loss}
\end{equation}

where $\lambda_{\mathrm{EMG}}$ controls
the contribution of EMG supervision.

\textbf{Action learning.}
We retain the original flow-matching
objective of the VLA backbone.
During training, demonstrated action
chunks are interpolated with Gaussian
noise at randomly sampled flow times.
The action expert learns to predict
the corresponding velocity using
mean-squared error (MSE) across
action dimensions and horizon positions.
This enables the policy to learn
action generation while preserving
the backbone's original training objective.

\textbf{Learning from EMG.}
Alongside action prediction, an auxiliary
head predicts the EMG representation
aligned with the starting action of
each chunk. As described in
Sec.~\ref{app:aux_head}, the prediction
is obtained from the pooled final-layer
VLM prefix features.

We consider two alternative supervision
targets: the sensor-wise activation
vector $\mathbf{y}_t^{\mathrm{act}}
\in\mathbb{R}^{8}$ and the learned
EMG embedding $\mathbf{z}_t^{\mathrm{EMG}}
\in\mathbb{R}^{128}$.
Accordingly, the auxiliary head predicts
$\hat{\mathbf{y}}_t^{\mathrm{act}}$
or $\hat{\mathbf{z}}_t^{\mathrm{EMG}}$,
respectively. Each variant is trained
separately using mean-squared error (MSE).

For either representation, we express
the auxiliary loss as

\begin{equation}
    \mathcal{L}_{\mathrm{EMG}}
    =
    \frac{1}{B}
    \sum_{i=1}^{B}
    m_i
    \frac{1}{d_y}
    \left\|
    \hat{\mathbf{y}}_i-\mathbf{y}_i
    \right\|_2^2,
    \label{eq:app_emg_loss}
\end{equation}

where $(\hat{\mathbf{y}}_i,\mathbf{y}_i)$
denotes either the predicted and target
sensor activations or the predicted and
target EMG embeddings. Here $B$ is the batch size, $d_y$
is the target dimension, and $m_i$
indicates whether the EMG target is
valid. 
The loss is averaged over the full
batch.

The two objectives influence different
parts of the network. While the action
loss trains the action expert and
shared VLM parameters, the EMG loss
updates the auxiliary head and 
VLM parameters.
In this way, muscle activity provides
an additional learning signal.

\textbf{Training configuration.}
For our primary real-world experiments,
we set $\lambda_{\mathrm{EMG}}=0.1$
and use AdamW with
$\beta_1=0.9$, $\beta_2=0.95$,
$\epsilon=10^{-8}$, and weight decay
$10^{-10}$. Gradients are clipped
to a maximum global norm of $1.0$.
The learning rate increases over
1,000 warmup steps to $2.5\times10^{-5}$,
then follows a cosine schedule reaching
$2.5\times10^{-6}$ at step 30,000.
Training is configured for 40,000 steps
with a global batch size of 32 and
an action horizon of $H=50$.

\textbf{Inference.}
EMG serves exclusively as a training
signal. Once training is complete,
the auxiliary prediction head is no
longer needed. The policy operates
using its standard visual observations,
robot state, and language instructions,
without requiring EMG measurements.

\section{Additional EMG-Action Association Controls}
\label{app:emg_controls}

Our main experiments demonstrate the
benefits of EMG supervision for robot
manipulation. We further investigate
whether the recorded muscle activity
exhibits meaningful associations with
robot actions beyond those expected
from randomized signals. Unlike our
MI-based temporal analysis, this experiment
examines the relationship between overall
muscle activation and commanded robot
movement using randomized EMG controls.
This provides complementary evidence
about the quality of the recorded signals.

\subsection{Diagnostic Data and Preprocessing}
\label{app:control_setup}

We analyze 400 provisionally paired
simulated-teleoperation demonstrations
from 10 participants. 
After offline filtering, we remove noisy
or near-constant signal intervals with
a one-second guard. We divide the
remaining signals into non-overlapping
100~ms bins, computing log-transformed,
within-trial standardized RMS activation.
The median across at least four valid
EMG channels represents overall muscle
activity, while the magnitude of three
normalized translation-command components
summarizes robot actions.

Each bin requires at least 95\% sample
coverage, and each demonstration requires
60 valid paired bins and at least 50\%
coverage of action-containing bins.
We compute Spearman correlation within
each demonstration and report the median
across demonstrations, ensuring equal
weight for each trajectory.

\subsection{Random-Signal Controls and Association Results}
\label{app:surrogates}

Having measured the association between
muscle activity and robot actions, we next
examine whether similar correlations could
arise from randomized signals. We construct
three complementary controls that
progressively preserve more properties
of the original EMG-derived activity,
while keeping robot actions unchanged.

\textbf{Gaussian noise.}
We first generate random signals matched
to the mean and standard deviation of
the measured activity. This provides
a simple reference without preserving
the original temporal structure.

\textbf{Fourier phase randomization.}
Next, we randomize the signal's Fourier
phases while preserving its frequency
amplitudes. This approximately retains
its autocorrelation structure but
disrupts the original temporal alignment
with robot actions.

\textbf{IAAFT surrogates.}
Finally, we use iterative amplitude-adjusted
Fourier transform (IAAFT) surrogates,
which preserve the empirical activity
distribution while approximately matching
the original frequency spectrum.
This provides a more constrained
comparison than Gaussian noise.

We generate 999 repetitions for each
control, constructing spectral surrogates
within continuous valid signal segments.
For each repetition, correlations are
computed per demonstration and summarized
using the median across demonstrations,
following the procedure in Sec.~\ref{app:control_setup}.

\begin{table}[t]
    \centering
    \caption{
        EMG-derived activity--action correlations
        compared with randomized signal controls.
        Spearman correlations are computed between
        overall EMG activity and robot translation
        command magnitude across 400 demonstrations.
        Surrogate values summarize 999 repetitions.
    }
    \label{tab:emg_surrogates}
    \footnotesize
    \setlength{\tabcolsep}{5pt}
    \renewcommand{\arraystretch}{1.15}
    \begin{tabular}{lcc}
        \toprule
        \textbf{Control Method} &
        \shortstack{\textbf{Measured}\\
                    \textbf{Activity--Action} $\boldsymbol{\rho}$} &
        \shortstack{\textbf{Surrogate}\\
                    \textbf{Activity--Action} $\boldsymbol{\rho}$} \\
        \midrule
        Gaussian noise          & \textbf{0.2964} & 0.0013 \\
        Fourier randomization   & \textbf{0.2964} & 0.0041 \\
        IAAFT                   & \textbf{0.2964} & 0.0021 \\
        \bottomrule
    \end{tabular}
\end{table}


\begin{figure}[t]
    \centering
    \includegraphics[
        width=\columnwidth,
        height=0.25\textheight,
        keepaspectratio
    ]{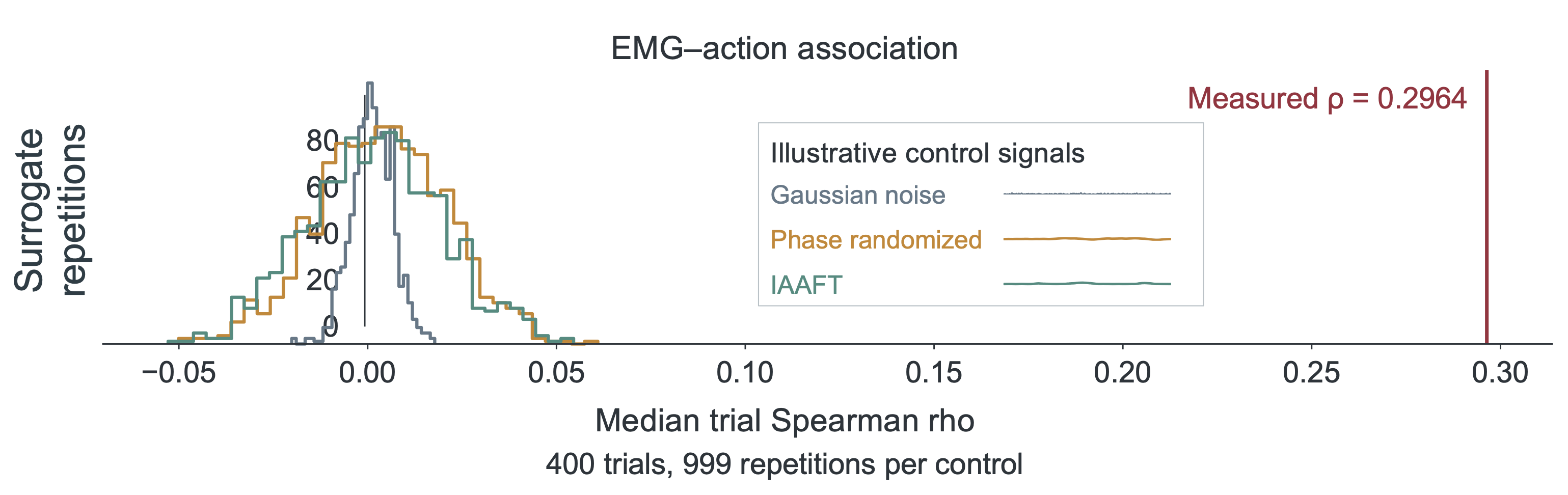}
    \caption{
        \textbf{EMG--action association under randomized controls.}
        Comparison of measured EMG-derived
        activity--action correlations with
        distributions obtained from randomized
        surrogate signals. Gaussian noise,
        Fourier phase randomization, and IAAFT
        provide complementary controls with
        different preserved signal properties.
        The measured association exceeds those
        obtained from the randomized controls,
        indicating temporal structure not
        reproduced by these surrogates.
    }
    \vspace{-4mm}
    \label{fig:association_controls}
\end{figure}

As shown in Table~\ref{tab:emg_surrogates}
and Fig.~\ref{fig:association_controls},
measured EMG-derived activity exhibits a
positive association with robot movement
commands ($\rho=0.2964$), whereas the
randomized controls yield correlations
near zero. The surrogate distributions
further illustrate that the observed
association is not reproduced by these
randomized signals.

\subsection{Interpretations}
\label{app:control_limitations}

The surrogate experiments provide an
additional perspective on the relationship
between human muscle activity and robot
actions. The measured EMG-action association
is substantially stronger than those
obtained from randomized signals, even
when the controls preserve aspects of
the original signal's frequency contents
and amplitude distribution. This suggests
that the temporal structure of recorded
muscle activity contains information
associated with commanded robot movement.

\section{Additional Experimental Protocols}
\label{app:experimental_protocols}

\subsection{Camera Occlusion Evaluation}
\label{app:camera_occlusion}

To investigate whether EMG-supervised
policies remain effective with incomplete
visual observations, we evaluate their
robustness to camera-view occlusion.
Rather than removing all visual input,
we selectively remove one camera view
while retaining the remaining observations,
robot state, and language instruction.

For each manipulation task, we conduct
10 occluded evaluation rollouts per policy,
divided equally between two conditions:
five rollouts with the center camera
removed and five with the left camera
removed. This allows us to examine
performance when different views of
the manipulation scene are unavailable,
while the remaining camera inputs
continue to provide visual information.

We additionally evaluate each policy
under clean observations using 10 rollouts
per task. All conditions use the same
task-specific success criteria, and
the occluded success rate is aggregated
across the two camera-removal conditions.

This evaluation is conducted separately
from the primary nine-task experiment,
which uses 50 rollouts per task.
Consequently, clean-condition success
rates may differ between the two studies.
Importantly, the EMG-supervised policy
receives no EMG measurements
during evaluation, allowing us to assess
whether its learned representations remain
useful when visual information is missing.

\subsection{Simulation Evaluation and Visual Perturbations}
\label{app:simulation_protocol}

We evaluate the baseline and EMG-supervised
policies trained on our human-teleoperated
MuJoCo dataset using robot rollouts under
clean and visually perturbed conditions.
Performance is measured through task
completion on held-out simulation tasks
and the LIBERO Spatial benchmark, rather
than predictions on recorded demonstrations.

The perturbed setting introduces visual
variation to examine whether the benefits
of EMG supervision extend beyond clean
observations. Both policies are evaluated
using the same task-success criteria
within each setting, with no EMG
measurements required at inference.
These evaluations complement the
real-world experiments by examining
EMG-supervised policy learning in
simulated manipulation environments.

\section{Additional Visual Attention Analysis}
\label{app:attention_analysis}

Beyond improvements in manipulation success,
we investigate how EMG supervision influences
the visual representations learned by the VLA.
Since our auxiliary objective directly
supervises the VLM's multimodal features,
we examine whether these representations
lead to different attention patterns
during robot manipulation.

Specifically, we investigate two questions:
where does the policy direct its visual
attention, and how concentrated is that
attention? We compare Robo-EMG against
the baseline using matched observations
from a real-world drawer-opening episode.

\subsection{Attention Around Interaction Regions}
\label{app:attention_visualization}

We visualize attention from action-token
queries to image-patch keys, as well as
instruction-token queries to image-patch
keys. Attention weights are averaged over
eight heads, normalized within each
camera view, and overlaid on the
corresponding RGB observations.

\begin{figure*}[h]
    \centering
    \includegraphics[
        width=\textwidth,
        keepaspectratio
    ]{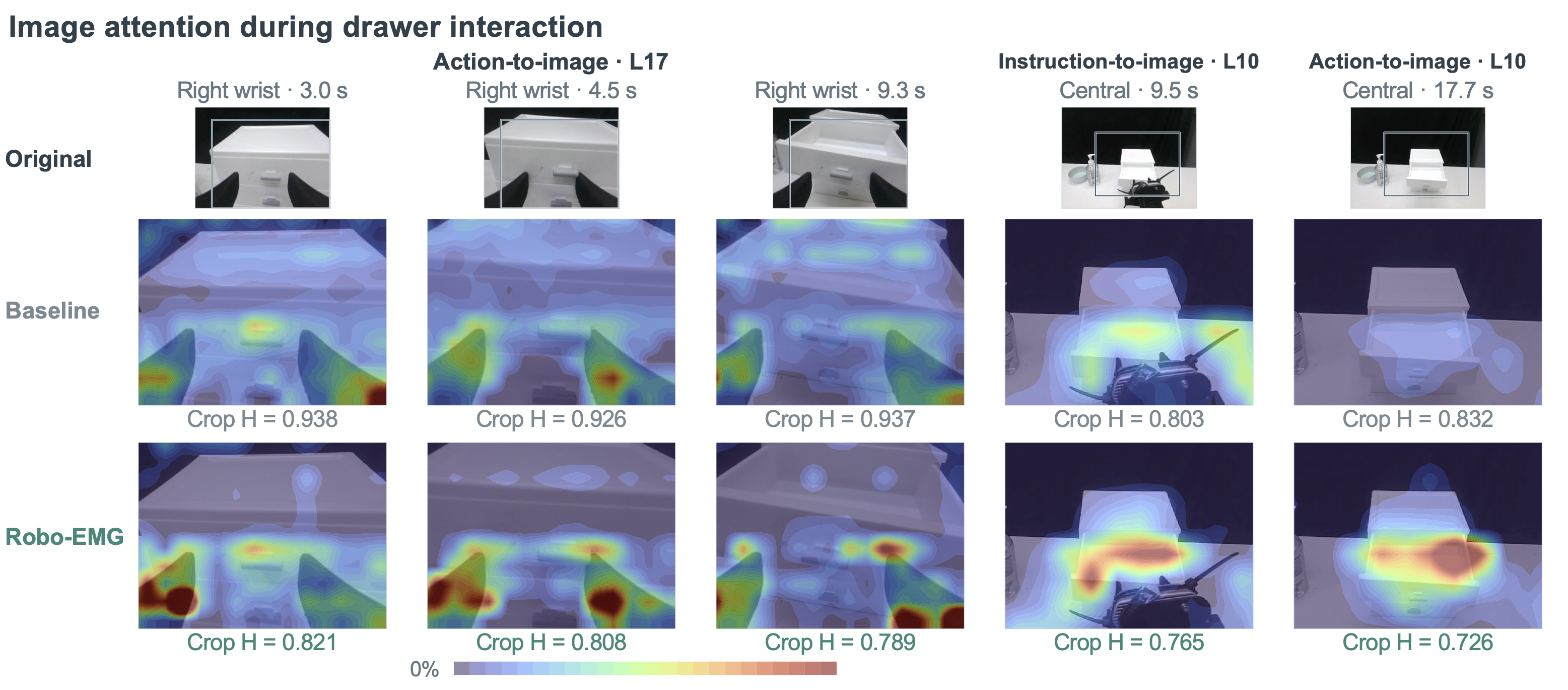}
    \caption{
        \textbf{Visual attention during drawer manipulation.}
        Matched observations and attention maps
        from the baseline and Robo-EMG.
        The visualizations include action-to-image
        attention at VLM layer 17 and
        instruction-to-image and action-to-image
        attention at layer 10.
        Robo-EMG exhibits stronger attention
        around manipulation-relevant regions,
        including the drawer handle and
        gripper--drawer interaction areas,
        whereas the baseline distributes
        attention more broadly across the scene.
        Spatial attention is averaged across
        eight heads and normalized within
        each camera view.
    }
    \label{fig:robo_emg_attention_visual}
\end{figure*}

Figure~\ref{fig:robo_emg_attention_visual}
shows representative comparisons at
different stages of drawer manipulation.
A clear distinction emerges between
the two policies. The baseline often
distributes attention broadly across
the drawer, surrounding surfaces, and
other scene regions. In contrast,
Robo-EMG exhibits stronger, more localized
attention around interaction-relevant
regions, particularly the drawer handle
and areas where the robot interacts
with the drawer.

This difference is visible across
multiple timesteps and attention-query
configurations. In the wrist-camera
examples, Robo-EMG emphasizes regions
near the gripper--drawer interaction,
while the baseline retains more diffuse
responses across the image. Similarly,
the central-camera examples show
more concentrated attention around
the drawer interaction region under
EMG supervision.

These visualizations suggest that
training with human muscle activity
encourages the VLM to allocate attention
more selectively during manipulation,
despite receiving no EMG measurements
at inference.

\subsection{Spatial Attention Entropy}
\label{app:attention_entropy}

While the visualizations suggest that
Robo-EMG focuses more strongly on
interaction-relevant regions, we further
examine whether its attention is
consistently less diffuse over time.
We quantify spatial concentration using
normalized attention entropy.

For a normalized spatial attention
distribution over $N$ image locations,
we compute

\begin{equation}
    H_{\mathrm{spatial}}
    =
    -\frac{1}{\log N}
    \sum_{i=1}^{N} p_i \log p_i,
    \label{eq:attention_entropy}
\end{equation}

where $p_i$ denotes the normalized
attention weight at image location $i$.
Higher entropy indicates more diffuse
attention across the image, whereas
lower entropy reflects greater spatial
concentration.

\begin{figure}[h]
    \centering
    \includegraphics[
        width=\columnwidth,
        keepaspectratio
    ]{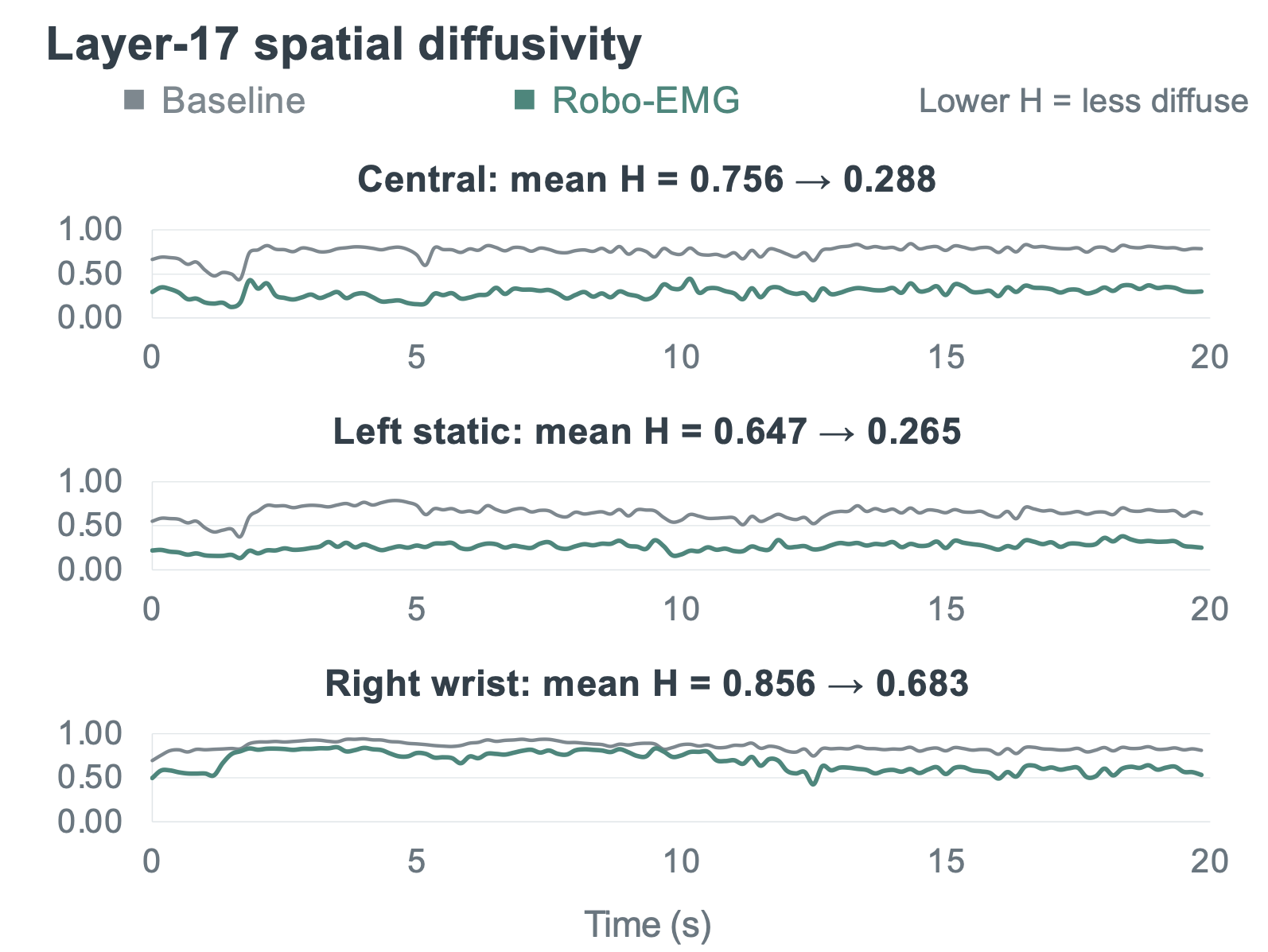}
    \caption{
        \textbf{Spatial attention diffusivity
        during drawer manipulation.}
        Normalized entropy of layer-17
        attention across 120 matched frames
        per camera view.
        Robo-EMG exhibits consistently lower
        entropy than the baseline, with mean
        values decreasing from
        0.756 to 0.288 (central),
        0.647 to 0.265 (left),
        and 0.856 to 0.683 (right wrist).
        Lower entropy indicates more
        concentrated spatial attention.
    }
    \label{fig:robo_emg_attention_diffusivity}
\end{figure}

We evaluate head-averaged spatial
attention from VLM layer 17 using
120 matched frames per camera view
from the same drawer-opening episode.
As shown in
Fig.~\ref{fig:robo_emg_attention_diffusivity},
Robo-EMG exhibits lower attention
entropy across all three cameras.
Mean entropy decreases from
0.756 to 0.288 in the central view,
0.647 to 0.265 in the left static view,
and 0.856 to 0.683 in the right wrist
view. Averaged across views, entropy
decreases from 0.753 for the baseline
to 0.412 for Robo-EMG.

These reductions complement the
qualitative attention maps: whereas
the baseline frequently spreads
attention across large portions
of the scene, Robo-EMG maintains
more spatially concentrated responses.
The effect is particularly pronounced
in the central and left static views.

\textbf{What can we learn from the attention analysis?}

Together, these analyses provide
complementary evidence that EMG
supervision influences the VLM's
visual representations. The qualitative
examples reveal stronger attention
around manipulation-relevant regions,
while the entropy measurements confirm
reduced spatial diffusivity throughout
the matched episode.

The analyses provide some evidence for the improved performance of the Robo-EMG model, and is in line with a lot of work which shows focused attention provides better results than diffused attention. Since this is a 'possible' explanation to the results shown in the paper and for the improved task success, we therefore interpret these findings
as supporting diagnostic evidence
of how EMG-supervised representation
learning changes visual attention.

\section{Dataset Task Inventory and Release}
\label{app:dataset_release}

To provide a more complete overview of
the Robo-EMG dataset, we list the 15
real-world manipulation tasks in
Table~\ref{tab:real_world_task_list}.
These tasks cover diverse manipulation
behaviors, including grasping, object
placement, stacking, and drawer interactions.
Each task contains 300 human-teleoperated
demonstrations paired with recorded EMG,
yielding 4,500 real-world trajectories.



\begin{table}[t]
    \centering
    \caption{%
        \textbf{Real-world Robo-EMG task inventory.}
        Fifteen single-arm manipulation tasks collected using ALOHA,
        with 300 human-teleoperated demonstrations per task.
    }
    \label{tab:real_world_task_list}

    \footnotesize
    \setlength{\tabcolsep}{3pt}
    \renewcommand{\arraystretch}{1.12}

    \makeatletter
    \setlength{\TX@delta}{0.1pt}
    \makeatother

    \providecommand{\taskicon}[1]{}
    \renewcommand{\taskicon}[1]{%
        \makebox[1.8em][l]{%
            \raisebox{-0.20\height}{%
                \includegraphics[
                    width=1.4em,
                    height=1.4em,
                    keepaspectratio
                ]{sec/figures/task_icons/#1.pdf}%
            }%
        }%
        \ignorespaces
    }

    \begin{tabularx}{0.97\columnwidth}{%
        @{}
        >{\centering\arraybackslash}p{0.07\columnwidth}
        >{\raggedright\arraybackslash}X
        @{}
    }
        \toprule
        \rowcolor{RoboBlue}[0pt][0pt]
        \textbf{\#} & \textbf{Manipulation Task} \\
        \midrule

        1 & \taskicon{cups}
            Stack three cups \\

        \rowcolor{RoboBlue}[0pt][0pt]
        2 & \taskicon{syringe}
            Place syringe on hand \\

        3 & \taskicon{tennis_can}
            Place tennis ball in can \\

        \rowcolor{RoboBlue}[0pt][0pt]
        4 & \taskicon{food_pan}
            Place food item on pan \\

        5 & \taskicon{drawer}
            Open top drawer \\

        \rowcolor{RoboBlue}[0pt][0pt]
        6 & \taskicon{drawer_cube}
            Open drawer and place cube \\

        7 & \taskicon{ball_transfer}
            Transfer tennis ball between containers \\

        \rowcolor{RoboBlue}[0pt][0pt]
        8 & \taskicon{sugar_cup}
            Place sugar cube in cup \\

        9 & \taskicon{cup_stand}
            Place cup in cup stand \\

        \rowcolor{RoboBlue}[0pt][0pt]
        10 & \taskicon{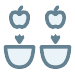}
             Sort red and green fruits into matching bowls \\

        11 & \taskicon{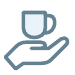}
             Hand cup to patient; release after secure grasp \\

        \rowcolor{RoboBlue}[0pt][0pt]
        12 & \taskicon{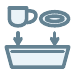}
             Place cup and plate into container sequentially \\

        13 & \taskicon{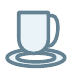}
             Place cup onto plate \\

        \rowcolor{RoboBlue}[0pt][0pt]
        14 & \taskicon{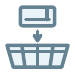}
             Place towels into basket one by one \\

        15 & \taskicon{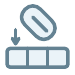}
             Place red pill in pill case \\

        \bottomrule
    \end{tabularx}
\end{table}

\noindent\textbf{Dataset release.}
We plan to publicly release Robo-EMG
and its accompanying code following
acceptance. Demonstrations will be
converted into the LeRobot format,
containing robot observations, states,
actions, task annotations, and temporally
aligned human EMG recordings.

In addition, we will separately release
the original raw eight-channel EMG signals
with their recording timestamps. This
allows researchers to explore alternative
signal-processing methods, investigate
EMG--action temporal relationships, and
develop new physiological representations
beyond those examined in our work.

\section{Limitations}
\label{app:limitations}

\textbf{EMG representations and generalization.}
Our framework currently uses pretrained,
task-specific EMG encoders that remain
frozen during VLA training. While effective, this adds an additional overhead on overall training time (since the encoder has to be trained before). Developing more
universal and transferable EMG representations
could make physiological supervision
more scalable and adaptable.

\textbf{Task-dependent benefits and alternative uses.}
Although EMG supervision improves performance
across our evaluated tasks, the magnitude
of improvement varies. Further analysis is
needed to understand which manipulation
behaviors benefit most from physiological
information. Moreover, while Robo-EMG uses
EMG exclusively for training supervision,
the synchronized recordings could also
support research into human intention
estimation and EMG-based robot control,
where reliable muscle-to-action mappings
remain an open challenge.

\textbf{Evaluation scope.}
Our experiments span real-world manipulation,
multiple VLA architectures, and simulated
environments, but broader evaluation across
unseen teleoperators, robot embodiments,
longer tasks, and more challenging conditions
is needed to establish generalizability.
We hope the public release of Robo-EMG,
including raw EMG and aligned demonstrations,
will facilitate these investigations.

\section{Future Work}
\label{app:future_work}

Robo-EMG provides an initial foundation
for investigating how human muscle activity
can enhance robot learning. By collecting
synchronized EMG and robot demonstrations,
characterizing their temporal relationships,
and demonstrating the benefits of EMG-based
supervision for VLA policies, our work
opens several promising research directions.

Beyond the tasks explored here, EMG could
provide valuable information about human
motor behavior during grasping, object
transport, contact-rich manipulation,
and fine-grained object interactions.
For example, muscle activation patterns
may reveal aspects of grasp preparation,
effort, and movement intention that are
not directly observable from robot
trajectories alone. Further investigation
could establish when such information
is most useful and how it can be
incorporated into different policy
learning frameworks.

The Robo-EMG dataset also provides a
starting point for studying more general
EMG-action relationships, developing
transferable muscle representations,
and combining EMG with complementary
human signals. By releasing both temporally
aligned robot demonstrations and raw
EMG recordings, we hope to enable
new approaches to human-informed robot
learning and encourage broader exploration
of physiological supervision across
robotic manipulation tasks.

\section{LLM Usage in Appendix}
ChatGPT was used to improve the readability
and organization of the Appendix, including
suggestions for the ordering of topics but 
without generating the underlying
content. Generative AI was also used to
create an initial layout for Figure~6 (initial blocks were created),
which was subsequently refined manually,
with its components replaced by original
experimental results. For Figure~9,
ChatGPT assisted in writing Python code
to visualize experimental
values over time. All tables, experimental
results, and final figure modifications
were produced and verified by the authors.

\end{document}